\documentclass{article}

\PassOptionsToPackage{numbers, compress}{natbib}

\usepackage[preprint]{neurips_2026}

\usepackage[utf8]{inputenc} 
\usepackage[T1]{fontenc}    
\usepackage{hyperref}       
\usepackage{url}            
\usepackage{booktabs}       
\usepackage{amsfonts}       
\usepackage{nicefrac}       
\usepackage{microtype}      
\usepackage{xcolor}         
\usepackage{scalerel}
\usepackage{tikz}
\usepackage{wrapfig}

\usepackage{graphicx}
\usepackage{doi}
\usepackage{caption}
\usepackage{subcaption}
\usepackage{amsmath}
\usepackage{amsthm}
\usepackage{verbatim}
\usepackage{multirow}

\title{
Subgraph Concept Networks: Concept Levels in Graph Classification
}

\author{
Lucie Charlotte Magister \\
  University of Cambridge\\
  Cambridge, UK\\
  \texttt{lcm67@cam.ac.uk} \\
  \And
  Alexander Norcliffe\thanks{Equal contribution} \\
  University of Cambridge\\
 Cambridge, UK\\
  \texttt{alin2@cam.ac.uk} \\
  \And 
  Iulia Duta\footnotemark[1] \\
  University of Cambridge\\
  Cambridge, UK\\
  \texttt{id366@cam.ac.uk} \\
  \And
  Pietro Li\`{o} \\
  University of Cambridge\\
  Cambridge, UK\\
  \texttt{pl219@cam.ac.uk} \\
}

\begin{document}

\maketitle

\begin{abstract}
The reasoning process of Graph Neural Networks is complex and considered opaque, limiting trust in their predictions. To alleviate this issue, prior work has proposed concept-based explanations, extracted from clusters in the model's node embeddings. However, a limitation of concept-based explanations is that they only explain the node embedding space and are obscured by pooling in graph classification. To mitigate this issue and provide a deeper level of understanding, we propose the Subgraph Concept Network. The Subgraph Concept Network is the first graph neural network architecture that distils subgraph and graph-level concepts. It achieves this by performing soft clustering on node concept embeddings to derive subgraph and graph-level concepts. Our results show that the Subgraph Concept Network allows to obtain competitive model accuracy, while discovering meaningful concepts at different levels of the network.
\end{abstract}

\section{Introduction}

Graph Neural Networks (GNN, \citep{scarselli2008graph}) suffer from an opaque reasoning process, which has inspired a number of works on increasing their interpretability for improved human trust \citep{corso2024graph, ji2025comprehensive, graphxai_survey}. A subset of works has focused specifically on extracting concept-based explanations \citep{magister2021gcexplainer, magister2023concept, azzolin2023globalexplainabilitygnnslogic}, because they provide global explanations easy to interpret by a human by extracting repeated graph patterns. These patterns are extracted from clusters formed in the GNN latent space by nodes with neighbourhoods. However, a common challenge observed by the Graph Concept Explainer (GCExplainer, \citep{magister2021gcexplainer}) and the Concept Distillation Module (CDM, \citep{magister2023concept}) is that no meaningful structure is observed in the latent space after pooling. This can be attributed to the use of mean pooling, which dilutes informative node-level signals by aggregating over many nodes. For a deeper understanding of graph classification, we therefore require an architecture that extracts subgraph and graph-level concepts after pooling. 

\begin{figure}
\centerline{\includegraphics[width=1\textwidth, trim=0.5cm 0cm 3.8cm 0cm, clip]{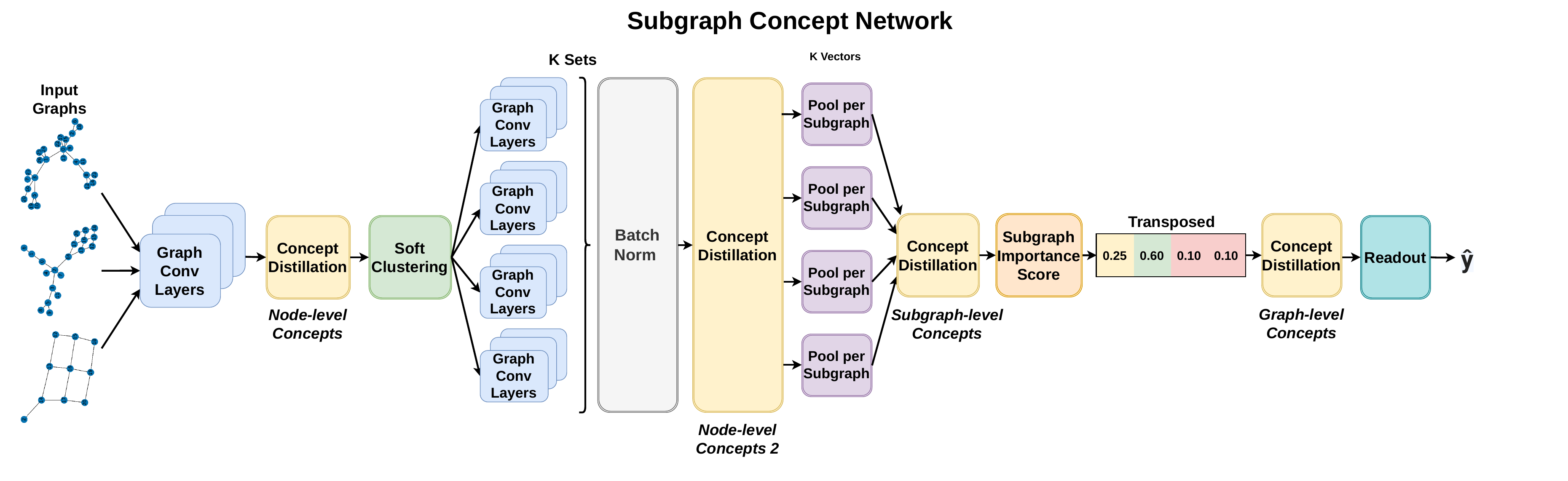}}
\caption{The SCN first embed the nodes of the input graph, after which we distil the first set of node-level concepts. We then perform soft clustering on the concept embeddings to extract $K$ subgraphs, which we further embed, before distilling a second set of node concepts at the subgraph-level. Finally, we perform a weighted pooling of the nodes in each subgraph and distil subgraph-level concepts, before obtaining a subgraph importance score for each of the pooled subgraph concept representations. Finally, we obtain a graph-level concept from these importance scores, which we use for prediction.}
\label{visual_abstract}
\end{figure}

To this aim, we introduce the Subgraph Concept Network (SCN). The SCN is the first GNN architecture that distils subgraph and graph-level concepts and uses these for graph classification. It achieves this by first embedding a graph to extract node-level concepts, on which it then performs soft clustering to extract subgraphs. SCN then uses these subgraph assignments to further embed the original graph per cluster, scaling the adjacency matrix based on the strength of the assignment of the nodes to the cluster. This allows to process the graph with focus on different subgraphs, encouraged through a custom loss function. An importance score for each of the pooled subgraph representations is then computed. The vector of the subgraph importance scores forms the graph-level concept, which is used for the final graph prediction. We experimentally show that the Subgraph Concept Network allows to (i) obtain competitive model accuracy, while (ii) discovering meaningful concepts at the node, subgraph and graph level. Figure \ref{visual_abstract} summarises the proposed architecture.

The remainder of this paper is structured as follows. In Section \ref{related_work}, we review the related literature and position our work, before presenting the SCN in Section \ref{method}. We then review the experimental setup in Section \ref{experiments} and present the results in Section \ref{results}. Finally, we conclude the paper with a discussion and conclusion in Sections \ref{discussion} and \ref{conclusion}.

\section{Related Work}
\label{related_work}

\paragraph{Interpretable Graph Neural Networks} A number of works focus on making GNNs more interpretable by extracting salient subgraph structures \citep{ji2025comprehensive}. The existing works can broadly be categorized into two main approaches: (1) prototypical-based methods \citep{zhang2022protgnn, seo2023interpretable, ragno2024, chen2019, wang2021tesnet, wang2024unveiling} and (2) information bottleneck–based methods. \citep{wu2020graph, yu2021recognizing, yu2022improving, lee2025pre}. Prototype-based methods can be regarded the closest to concept-based methods, wherefore, we focus on a review of them. An early example of a prototype-based approach is ProtGNN \citep{zhang2022protgnn}, which learns prototypical graph patterns using MCTS, and then uses these for classification. Similarly, the Prototype-based Graph Information Bottleneck (PGIB, \citep{seo2023interpretable}) learns prototypical graph structures, however, using the information bottleneck framework \citep{tishby2000information}. Also relying on prototypes, \citet{ragno2024} adapt the Prototypical Part Network (ProtoPNet, \citep{chen2019}) and the transparent embedding space (TesNet, \citep{wang2021tesnet}) method from the image to the graph domain. Similarly, Global Interactive Pattern (GIP, \citep{wang2024unveiling}) learning also relies on graph prototypes. The GIP method first clusters nodes using a constrained graph clustering module, after which it then matches coarsened global interactive instances with a batch of self-interpretable graph prototypes. The use of clustering makes the GIP method the most comparable method to SCNs, however, it requires setting a number of prototypes to learn per class. This is a common drawback of prototype-based methods. Nevertheless, the GIP method also introduces custom loss terms for more interpretable clustering, informing the loss terms used to train the SCN. While learned prototypes can be seen as a version of concepts, the aforementioned methods do not operate at different concept levels like the SCN. 

\paragraph{Graph Pooling} The problem of graph pooling is closely related to distilling subgraph and graph-level concepts in GNNs. In CDM, we found that simple average pooling obscures the latent space, as the aggregation of all nodes dampens the signal of important ones. Moreover, it does not consider the interplay between node concepts. More advanced graph pooling techniques, such as DiffPool \citep{diffpool} and StructPool \citep{Yuan2020StructPool}, iteratively coarsen the graph, yielding more interpretable latent spaces. Similarly, the Tree-like Interpretable Framework (TIF, \citep{yang2025gnns}), relies on iterative graph coarsening to transform the a GNN into a hierarchical tree, where each level represents a coarsened graph. Our work draws inspiration from DiffPool and TIF in regard to the soft clustering of nodes. However, we then use these clusters for subgraph re-embedding, rather than graph coarsening. Hierarchical Explainable Latent Pooling (HELP) \citep{2023everybodyneedslittlehelp} also learns to cluster connected node embeddings and pools these. While HELP also considers concepts for pooling, it does not consider the presence of multiple subgraph concepts but also rather relies on coarsening. Furthermore, while the SCN draws inspiration from DiffPool and HELP, it is not a pooling method itself. Instead, the SCN is a GNN architecture designed to distil multiple subgraph and graph-level concepts for graph classification.

\section{Subgraph Concept Network}
\label{method}

We propose the Subgraph Concept Network (SCN), the first GNN for graph classification designed to identify subgraph and graph-level concepts for prediction. The SCN operates in three steps. First, the SCN embeds the graph as node concepts and performs soft clustering to partition it into multiple subgraphs, which are then further embedded. Second, the SCN pools the node concept embeddings of each of the identified subgraphs to obtain a subgraph concept embedding, for which it predicts an importance score. Lastly, the SCN constructs a graph concept embedding based on these importance scores and uses it for prediction. In addition to this architecture, we introduce a custom loss function to optimise the SCN during training, inspired by the custom loss function used by the GIP method \citep{wang2024unveiling}.

\paragraph{Node Concept Distillation} The first step of the SCN focuses on distilling node-level concepts, as proposed in CDM. Specifically, the SCN first embeds a graph $\mathcal{G}$ using graph convolutions to obtain an embedding $h_i$ of size $s$, where $s$ is the size of the node-level encoding vector: $h^l_i = \mathrm{GraphConv}^l(h^{l-1}_i, \mathcal{A_G})$, where $h^l_i$ is the embedding of node $i$ after $l$ $\mathrm{GraphConv}$ layers and $\mathcal{A_G}$ is the adjacency matrix of graph $\mathcal{G}$. As proposed in CDM \citep{magister2023concept}, we then extract node-level concepts by applying a normalized softmax activation function on the node embeddings $h^l_i$ to obtain a fuzzy concept encoding  $\mathbf{q}_i$ of size $s$ for each node. Given these node-level concepts, we then perform subgraph extraction.

\paragraph{Subgraph Concept Distillation} The second step of the SCN focuses on distilling node-level concepts at the graph level into node concepts at the subgraph level, and finally subgraph-level concepts. Inspired by DiffPool \citep{diffpool}, the SCN identifies subgraphs through soft clustering. We first perform a normalized graph convolution to project all node concepts across graphs into the same embedding space for clustering. Inspired by DiffPool \citep{diffpool}, we then identify subgraphs by applying another softmax activation function to obtain a soft cluster assignment $c_i \in [0, 1]^{K}$: $c_{i} = \frac{exp(h^{l+1}_i)}{\sum^{k}_{u=1} exp(h^{l+1}_{iu})}$ where $K$ is the number of subgraphs to use. Similar to DiffPool \citep{diffpool} and HELP~\citep{2023everybodyneedslittlehelp}, we consider the nodes strongly assigned to a cluster as a subgraph. However, instead of pooling these nodes, we use the cluster assignment matrix to re-weight $h^{l+1}_i$  and $\mathcal{A_G}$, such that we can perform $K$ separate embeddings of the graph, focusing on the different subgraphs. We obtain these re-weighted node embeddings in the following way: $h_{ik} = h^l_{i} c_{ik}$, where $h_{ik}$ is the re-weighted node embedding $h^{l+1}_i$ based on the cluster assignment value $c_{ik}$ of node $i$ for cluster $k$. Note, that we do not embed the original node embeddings, but the already processed ones for improved gradient flow. We then compute the re-weighted adjacency matrices for each cluster $k$ as follows: $\mathcal{A}_k = (\sum^{k}_{m=1} c_{im} c_{jm}) \cdot \mathcal{A}_{ij}$, where $i$ and $j$ are indices for the corresponding cluster assignment value and adjacency matrix value of nodes $i$ and $j$, respectively.

We now have $K$ sets of scaled nodes and adjacency matrices, which we further embed using another $l$ graph convolutional layers, as if we where re-embedding the original graph. Note, that we use $K$ separate sets of graph convolutional layers for re-embedding to specialise on the different subgraphs. We then batch normalise the node embeddings across the subgraphs to project them into the same latent space and perform node-level concept distillation per subgraph as in CDM \citep{magister2023concept}. Next, we apply an approximate version of mean pooling to the node-level concept embeddings per subgraph to obtain the subgraph embeddings $h_k$. We use an approximate version of mean pooling, as we soft assign nodes to a cluster / subgraph. Therefore, we perform mean pooling weighted only by the strength of the assignment of the nodes in the subgraph: $h_{k} = \frac{\sum_{i=1}^{n} h_{ik}}{\sum^n_{i=1} c_{k} + \epsilon}$, where $n$ is the number of nodes in graph $\mathcal{G}$, $h_{ik}$ is the scaled embedding of node $i$ for subgraph $k$ and $c_k$ is the cluster assignment vector for subgraph $k$. Notice, that we add $\epsilon$ to the denominator to avoid division by 0 if some clusters remain unused. Overall, this leaves us with $K$ subgraph embeddings characterising the graph $\mathcal{G}$. We extract subgraph-level concepts from these embeddings by applying another normalized softmax activation function as in CDM \citep{magister2023concept}.

\paragraph{Graph Concept Distillation} Given the $K$ subgraph concepts per graph, we then construct a graph embedding. Taking inspiration from the Deep Concept Reasoner (DCR, \citep{dcr2023}), we use a neural model to determine the role of the subgraph concept. Specifically, we use a feed-forward network $\Theta : \mathbb{R} \rightarrow [0, 1]$ to predict an importance score per subgraph, which we then concatenate to obtain the graph embedding $h_\mathcal{G}$. We then distil graph-level concepts following CDM \citep{magister2023concept} again, before finally applying a simple readout function for prediction.

\paragraph{Custom Loss} We propose a novel loss function that balances minimising the cross-entropy loss $\mathcal{L}_{CE}$ for predictive accuracy with four additional loss terms for encouraging interpretable subgraph representations. The first additional loss term we introduce is the entropy loss $\mathcal{L}_{\text{entropy}}$, as also proposed in DiffPool \citep{diffpool}. The entropy loss focuses on encouraging nodes to be strongly assigned to a cluster. As we soft assign nodes to a cluster in our architecture, the entropy loss function is based on Shannon's entropy \citep{entropy}, measuring the uncertainty in a probability distribution. While we take inspiration from the entropy loss defined in DiffPool \citep{diffpool}, we make two important modifications. Firstly, we normalise the entropy loss in order to balance it with the other losses. Secondly, we mask invalid padding nodes in the graph, which are necessary for the adjacency matrix-based implementation of the SCN, but should not influence the loss computation or clustering results.

Similar to DiffPool \citep{diffpool}, we also encourage the nodes assigned to a cluster to be connected via the connectivity loss $\mathcal{L}_{\text{connectivity}}$. Specifically, DiffPool \citep{diffpool} uses the Frobenius norm of the difference between the adjacency matrix and the outer product of the cluster assignments to measure how well the soft cluster similarities reflect the global graph structure. In contrast, we opt for a connectivity loss that consists of three components: a positive similarity penatly, a negative similarity penalty and an isolation penalty. We draw inspiration for this three component penalty loss from the GIP method \citep{wang2024unveiling}. \citet{wang2024unveiling} proposes a three component penalty loss to target the problem of aligning a coarsened subgraph only with prototypes of one class. In contrast, we adapt the loss to assign connected nodes to the same cluster.

The positive similarity penalty $\mathcal{P}_{\text{pos\_sim}}$ penalises connected node pairs that have a dissimilar soft cluster assignment. We can achieve this by computing the dot product of the cluster assignments for each node pair $(i, j)$ where $A_{ij} = 1$, and then applying a hinge-like penalty, encouraging connected node pairs to have a minimum level of cluster assignment similarity. In contrast, the negative similarity penalty $\mathcal{P}_{\text{neg\_sim}}$ discourages disconnected nodes from being assigned to the same cluster. We achieve this by penalising disconnected node pairs by the similarity of their cluster assignments, computed by their inner product. This discourages disconnected nodes from being assigned to the same cluster, preventing mode collapse, where all nodes are assigned to the same cluster. Lastly, the isolation penalty penalises isolated nodes within a cluster assignment. We achieve this by summing the cluster similarity of a node $i$ and its neighbours in the graph. If the total cluster similarity falls below a certain threshold $\tau$, the node is considered isolated. We can then combine the three components into the connectivity loss $\mathcal{L}_{\text{connectivity}}$, scaling their contribution. Specifically, we prioritise avoiding isolated nodes.

In contrast to DiffPool \citep{diffpool}, we also introduce two further loss terms to accommodate our network design, of which one is the utilisation loss $\mathcal{L}_{\text{utilisation}}$. This loss term discourages the model from assigning all nodes to a single cluster. This problem was also observed by \citet{wang2024unveiling} and solved via an additional loss term. To achieve a balanced cluster usage, we first compute the average assignment of all nodes to each cluster to obtain a distribution reflecting the cluster utilisation. We then compute Shannon's entropy \citep{entropy} of this distribution to quantify how evenly assignments are spread across clusters. We then normalise the entropy by the maximum possible value, $log(K)$, to make sure the result lies between 0 and 1, like the other losses. We subtract the normalised entropy from one, so that lower loss values correspond to more uniform and desirable cluster usage, as for the other loss terms.

Lastly, we introduce the consistency loss $\mathcal{L}_{\text{consistency}}$, which encourages consistent clustering across graphs by promoting similar node embeddings across graphs to be assigned to the same cluster. This is important because we use a vector of subgraph importance scores for the final prediction and therefore want to encourage an ordering among subgraphs common across graph input instances. Specifically, we calculate the normalised Euclidean distance between the soft cluster assignments of each node pair in a batch of graphs. We weight these distances based on how similar the nodes are in the embedding space, using a scaled and normalised cosine similarity. This approach finds node pairs that are close in the embedding space across the batch of graphs. To train the model effectively, we combine these four custom losses with the cross entropy loss with individual scalars, which we will discuss in Section \ref{results}.

\paragraph{Quantifying subgraph extraction success}
\label{subgraph_scores}

Given the design of the SCN, we propose measuring the success of the subgraph extraction, which forms the basis for the subgraph concepts distilled. We can first measure the subgraph assignment strength, quantifying how strongly assigned nodes are to a subgraph via clustering. For this, we can simply average across the maximum assignment probability of a node to a cluster. The maximum probability at which a node can be assigned to a cluster is $1.0$. The lower bound on the subgraph assignment is based on the score taking the maximum probability value. When a node is assigned to each cluster with equal probability, the score would minimally be $\frac{1}{K}$. A value of $1$ is desirable here, as it indicates that all nodes are strongly assigned to a cluster. While each node is only counted towards one cluster, a drawback of this metric is that it does not quantify how distributed the assignment of the node is across the remaining clusters. In the worst case, all nodes are strongly assigned to one cluster.

Based on the drawback identified, we can compute the conditional subgraph assignment strength per cluster by only considering nodes with a probability of $c_{ik} \ge \frac{1}{K}$ to be assigned to a cluster, as any value below this threshold would be less than the probability of a node being assigned to all clusters equally. We can then compute the mean over these scores to obtain a measure of cluster utilization. If the mean approaches $\frac{1}{K}$, all nodes have been assigned to one cluster. 

Lastly, we can compute a subgraph consistency score by computing if node-level concept embeddings are consistently assigned to the same cluster. Specifically, we compare the cluster label assigned to a node of a specific concept to the cluster used most frequently by this concept and average over all node concepts across the graphs. For this, we define a node to be assigned to the cluster with the highest probability. This metric allows us to check how consistently nodes of a given concept are assigned to the same cluster across graphs. We can only compute this metric because of the notion of concepts in our network allowing to match nodes across graphs.

\section{Experiments}
\label{experiments}

We focus on answering the following research questions in our experiments: (i) how does the accuracy of the SCN compare to a standard graph-classification GNN?, (ii) is the extracted concept set complete with respect to the task? and (iii) are the unsupervised concepts identified by our model meaningful?

\paragraph{Metrics} We evaluate the performance and interpretability of SCN using three key metrics. In line with our first research question, we measure model performance via classification accuracy, comparing the generalisation error of SCN with equivalent standard GNNs. In line with our second and third research question, we evaluate model interpretability via \textit{concept completeness}. Recall, that concept completeness measures whether the extracted set of concepts is sufficient to describe the downstream task. Following \citet{yeh2020completeness} we use a decision tree~\cite{breiman1984classification} to predict the task label from the concept encoding associated with an input instance. In contrast to CDM \citep{magister2023concept}, we can compute the concept completeness score at different levels, namely the node, subgraph and graph level. Lastly, we propose and compute three different clustering scores to measure the performance of our subgraph extraction, described in Section \ref{subgraph_scores}. We do not compute concept purity as proposed in \cite{magister2021gcexplainer}, as the concept size of subgraph-level concepts increases significantly, making the graph edit distance of a subgraph infeasible to compute.

\paragraph{Datasets} The SCN is specifically developed for graph classification, wherefore, we only consider graph classification tasks. We consider six datasets in total, including four synthetic datasets and two real-world datasets. The four synthetic graph datasets are Grid, Grid-House, STARS and House-Colour, introduced by \citet{longa2022explaining}. These datasets encode ground truth motifs that an interpretable model should successfully recover. Inspired by the BA-Grid dataset for node classification introduced by \citet{ying2019gnnexplainer}, the Grid dataset for graph classification consists of 1000 BA graphs, where a 3-by-3 grid structure is attached to half of the graphs. Here, the positive class is assigned to the graphs with an attached grid structure. The Grid-House dataset is constructed in a similar manner, however, with a grid or house structure attached. Here, the positive class is assigned to graphs that have either a grid \underline{or} house structure attached, but not both. In contrast, the STARS benchmark consists of random graphs generated by the Erd\H{o}s--R\'enyi (ER) random graph model \citep{Erdos1984OnTE} with one to four star-shaped structures attached. Here, class 0 and 1 correspond to graphs with 1 and 2 star attached, respectively, while class 2 corresponds to graphs with 3 or 4 stars attached. Lastly, the House-Colour dataset is also based on the BA graph and the house motif. Different to the previously listed datasets, the House-Colour dataset includes node features. Specifically, the nodes in the base graph are randomly assigned the colours blue, green or red, represented as a one-hot vector. One to three house structures with a uniform colour are then attached to the base graph, where a blue house indicates the positive class and a green house the negative class, while the remaining houses have randomly coloured nodes. 

In contrast to the synthetic datasets, the two real-world datasets Mutagenicity and Reddit-Binary \citep{morris2020tudataset} are less structured. Mutagenicity \citep{morris2020tudataset} is a collection of graphs representing mutagenic and non-mutagenic molecules, which should be identified as such. While the dataset does not have ground truth motifs, \citet{ying2019gnnexplainer} suggest the ring structure and nitrogen dioxide compound as desirable motifs to recover. Reddit-Binary \citep{morris2020tudataset} is a collection of Reddit discussion threads, where nodes represent users and edges represent interactions. Here, the desirable motif to recover is a star-like structure, according to \citet{ying2019gnnexplainer}. We refer the reader to Appendix \ref{dataset_stats} for a summary of dataset statistics.  

\paragraph{Baselines and Setup} As the SCN is the first GNN with multiple concept levels, we compare the model to variants of a CGN, as they remain the only concept-based interpretable model architecture to the best of our knowledge. Specifically, we compare the SCN to a slightly modified vanilla CGN and a CGN with DiffPool \citep{diffpool}. We slightly modify the vanilla CGN to (1) obtain more concept levels and (2) ablate our design choices. Specifically, the vanilla CGN is composed of a number of embeddings layers, followed by a normalized softmax function for node-level concept extraction. These node-level concepts are then simply pooled using mean pooling. We then obtain graph level concepts by inserting another normalized softmax activation function, before applying a simple linear readout function. Note, that we do not use a LEN as the readout function here, because we want to keep the networks as comparable as possible. The CGN with DiffPool \citep{diffpool}, follows the same overall structure but inserts a DiffPool layer followed by a normalized softmax activation function on the coarsened embeddings before the mean pooling. This architectural change also allows us to extract subgraph-level concepts and validate the design choices of the SCN, which were inspired by DiffPool \citep{diffpool}. We also compare to the GIP \citep{wang2024unveiling} method, since we draw inspiration from it and it is the closest related work to ours. However, note that we cannot fully compare in the same manner, as the GIP method does not operate in the concept space. We refer the reader to Appendix \ref{model_stats} for a summary of the model training. Note, that we do not compare to post-hoc methods, such as GNNExplainer \citep{ying2019gnnexplainer} or GLGExplainer \citep{azzolin2023globalexplainabilitygnnslogic}, as we are focused on assessing advances in model interpretability, rather than explainability. We also do not compare our method to HELP \citep{2023everybodyneedslittlehelp}, as HELP is a pooling method for node-concepts. In contrast, SCN is a GNN architecture, focused on distilling multiple levels of concepts and using these for pooling and prediction.

\section{Results}
\label{results}

\paragraph{Subgraph concept networks are as accurate as concept graph networks}

In reference to Table \ref{fig:accuracy}, our results show that the SCN is as accurate as a vanilla CGN and a CGN using DiffPool \citep{diffpool}, as well as the GIP method \citep{wang2024unveiling}. Let us first examine the CGN-based models in comparison to the SCN. The SCN outperforms the CGN using mean pool and DiffPool on the Grid, Grid-House, Mutagenicity and Reddit-Binary datasets. It also achieves on par performance on the the STARS and House-Colour datasets, with less than 0.5 accuracy points difference. Most notably, the SCN outperforms the CGN variants on the Grid-House dataset, achieving an accuracy of 89.50\%, while the CGN with mean pooling and DiffPool only achieve an accuracy of 80.40\% and 50.00\%, respectively. Most surprisingly, the CGN using DiffPool appears to struggle with identifying the different combinations of grid and house structures attached to the base graph, not exceeding random accuracy. This indicates that the SCN performs better at identifying important subgraphs and using these for prediction. Comparing the SCN to the GIP method, the GIP method slightly outperforms the SCN on the four synthetic datasets. Overall, we can call the performance of the SCN and GIP method on par on the Grid, STARS and House-Colour datasets, with less than 1\% accuracy difference. However, the SCN only achieves an accuracy of 89.50\% on the Grid-House dataset, while the GIP method achieves a perfect accuracy of 100\%. In contrast, the SCN performs better on the real-world tasks than the GIP method, achieving an accuracy of 80.53\% and 91.23\% compared to an accuracy of 78.42\% and 88.89\% on the Mutagenicity and Reddit-Binary datasets, respectively.

\begin{table*}[ht]
\caption{Model accuracy for the Subgraph Concept Network (SCN) and two Concept Graph Networks (CGN) using mean pooling and DiffPool, as well as the GIP method \citep{wang2024unveiling}.}
\centering
\resizebox{0.95\textwidth}{!}{
\begin{tabular}{lllll}
\hline & \multicolumn{4}{c}{\textbf{\begin{tabular}[c]{@{}c@{}}Model Accuracy (\%)\end{tabular}}} \\
                       & \multicolumn{1}{c}{\textbf{SCN}}      & \multicolumn{1}{c}{\begin{tabular}[c]{@{}c@{}}\textbf{CGN}\\\textbf{w/ mean pool}\end{tabular}} & \multicolumn{1}{c}{\begin{tabular}[c]{@{}c@{}}\textbf{CGN}\\\textbf{w/ DiffPool}\end{tabular}} & \multicolumn{1}{c}{\textbf{GIP}} \\ \hline
\textbf{Grid}    & 99.40 (98.59, 100.00) & 98.6 (98.08, 99.12) & 98.05 (93.15, 100.00) & \textbf{100.00 (100.00, 100.00)}\\
\textbf{Grid-House}  & 89.50 (84.98, 94.02) & 80.40 (74.62, 86.18) & 50.00 (50.00, 50.00) & \textbf{100.00 (100.00, 100.00)} \\
\textbf{STARS} & 99.13 (98.58, 99.69) & 98.87 (98.17, 99.56) & 99.20 (98.73, 99.67) & \textbf{99.27 (98.92, 99.61)} \\
\textbf{House-Colour}    & 99.60 (98.49, 100.00) & 99.60 (99.08, 100.00) & 98.90
(97.79, 100.00) & \textbf{100.00 (100.00, 100.00)} \\
\hline
\textbf{Mutagenicity}  & \textbf{80.53 (79.46, 81.61)}  & 78.77 (77.87, 79.67) & 79.67 (77.12, 82.23) & 78.42 (77.45, 79.38)                         \\
\textbf{Reddit-Binary} & \textbf{91.23 (88.74, 93.73)} & 90.49 (89.02, 91.96) & 89.88 (88.54, 91.22) & 88.89 (85.07, 92.71)         \\
\hline
\end{tabular}%
}
\label{fig:accuracy}
\end{table*}

\paragraph{Subgraph concept networks discover a complete set of subgraph-level concepts} The SCN specifically introduces subgraph-level concepts for more interpretable graph prediction. In Table \ref{fig:completeness_accuracy2}, we collate subgraph-level concept completeness scores computed from different combinations of concept representations. First, we compute the average concept completeness per subgraph. This hovers at slightly above 50\% on each of the datasets. However, what this metric neglects is that some subgraphs may be common across graphs of different classes and that not all subgraphs discovered are important for the final prediction. For example, the Grid dataset differentiates classes by whether a grid structure is attached to the base graph. In this case, predicting the final target from just one subgraph, which would be the grid or base graph, would not be predictive. Due to the presence or combination of subgraphs being important, it makes more sense to concatenate the subgraph concept representations and predict the final target from this. This is also validated by the increase in the concept completeness score achieved. For example, the concept completeness score for Grid dataset improves from 55.92\% on the individual subgraph concept level to 61.60\% when we concatenate the subgraph concept embeddings. However, we achieve the best results, when we concatenate the subgraph concept embeddings and the importance scores of the subgraphs. For example, on the Grid-House dataset, we now achieve a concept completeness score of 97.60\%. The concept completeness score is comparable to the model accuracy of 99.40\%, which is deemed the upper bound for concept completeness \citep{magister2021gcexplainer}. We can compare the subgraph-level concept completeness score of the SCN only with that of the CGN using DiffPool \citep{diffpool}, as the DiffPool layer allows to extracts subgraph level representations. Comparing the individual concept completeness scores and the completeness scores computed on the concatenated subgraph representations, we observe the CGN using DiffPool to achieve higher concept completeness levels. For example, on the Grid dataset, the CGN with DiffPool achieves an individual concept completeness score of 81.63\% and a concatenated subgraph concepts completeness score of 83.80\%, outperforming the SCN. This could be attributed to the SCN learning the task based on subgraph importance scores, while the CGN using DiffPool ultimately still relies on a single graph representation obtained from hierarchical pooling. Moreover, when we examine the clustering of DiffPool, we find that it does not help interpretability. Examining a few cluster assignments for the Grid dataset, we observe that all nodes are assigned to one cluster with a probability of $\sim1$, with only a few nodes being assigned to both clusters with a probability of $\sim0.5$. This indicates that the subgraph concept embeddings are more like graph level embeddings. However, the interpretability of the network is not improved as no clear subgraphs are identified. In contrast, the SCN strongly assigns nodes to clusters across multiple clusters. Furthermore, the SCN achieves a higher concept completeness scores when we concatenate the concept and importance score representations. For brevity, we refer the reader to Appendix \ref{node_level_concept_completeness} for an analysis of node and graph-level concept completeness, and a comparison to the GIP method \citep{wang2024unveiling}.

\begin{table*}[t]
\caption{Subgraph-level concept completeness scores for the Subgraph Concept Network (SCN) and Concept Graph Network (CGN) with DiffPool \citep{diffpool}.}
\centering
\resizebox{\textwidth}{!}{
\begin{tabular}{llll|ll}
\hline  & \multicolumn{5}{c}{\textbf{\begin{tabular}[c]{@{}c@{}}Subgraph-level Concept Completeness (\%)\end{tabular}}} \\
& \multicolumn{3}{c}{\textbf{SCN}}          
& \multicolumn{2}{c}{\begin{tabular}[c]{@{}c@{}}\textbf{CGN} \textbf{w/ DiffPool}\end{tabular}} \\
& \multicolumn{1}{c}{\begin{tabular}[c]{@{}c@{}}\textbf{Individual}\\\textbf{Concepts}\end{tabular}}
& \multicolumn{1}{c}{\begin{tabular}[c]{@{}c@{}}\textbf{Concat.}\\\textbf{Concepts}\end{tabular}}
& \multicolumn{1}{c}{\begin{tabular}[c]{@{}c@{}}\textbf{Concat. Concepts}\\\textbf{w/ imp. scores}\end{tabular}}
& \multicolumn{1}{|c}{\begin{tabular}[c]{@{}c@{}}\textbf{Individual}\\\textbf{Concepts}\end{tabular}}
& \multicolumn{1}{c}{\begin{tabular}[c]{@{}c@{}}\textbf{Concat.}\\\textbf{Concepts}\end{tabular}} \\
\hline
\textbf{Grid} & 55.92 (51.98, 59.87) & 61.60 (51.42, 71.78) & \textbf{97.60 (94.37, 1.00)} & 81.63 (61.52, 100.00) & 83.80 (56.11, 100.00)\\
\textbf{Grid-House} & 51.98 (48.91, 55.04) & 57.40 (49.24, 65.56) & \textbf{82.30 (74.21, 90.39)} & 50.00 (50.00, 50.00) & 50.00 (50.00, 50.00) \\
\textbf{STARS} & 55.23 (42.41, 68.05) & 76.47 (57.79, 95.15) & \textbf{98.07 (97.39, 98.75)} & 66.60 (45.21, 87.99) & 83.67 (66.61, 100.00) \\
\textbf{House-Colour} & 48.00 (42.57, 53.43) & 50.40 (43.04, 57.76) & \textbf{98.30 (97.26, 99.34)} & 93.15 (80.02, 100.00) & 88.30 (60.64, 100.00)        \\
\hline
\textbf{Mutagenicity}  & 56.62 (54.58, 58.66) & 56.63 (54.63, 58.63) & \textbf{76.70 (75.70, 77.69)} & 75.73 (75.16, 76.30) & 76.70 (74.79, 78.61)               \\
\textbf{Reddit-Binary} & 56.78 (53.52, 60.05) & 59.44 (54.25, 64.64) & \textbf{88.64 (86.43, 90.85)} & 79.09 (70.44, 87.73) & 83.15 (76.15, 90.14)             \\
\hline
\end{tabular}%
}

\label{fig:completeness_accuracy2}
\end{table*}

\subsection{Study of the clustering behaviour}

We introduce four custom loss terms to encourage interpretable clustering in the model. To study the effect of these losses, we recommend computing 3 metrics quantifying different interpretability aspects of the clustering. For brevity, we will focus on cluster utilization and refer the reader to Appendix \ref{analysis_cluster_usage} for an analysis of the other metrics. The conditional subgraph assignment metric computes the strength of the nodes assigned to a cluster per cluster, where a node is deemed to be assigned to a cluster if its activation exceeds $\frac{1}{K}$, where $K$ is the number of clusters. Given this score, we can compute cluster utilization. Table \ref{fig:conditional_subgraph_assignment_strength} summarizes the cluster utilization for the SCN and CGN with DiffPool \citep{diffpool}. Overall, we find that the SCN exhibits better cluster utilization than the CGN using DiffPool. For example, the cluster utilization of the SCN exceeds $0.89$ on the Grid, Grid-House, STARS and House-Colour datasets across both the train and test slice, indicating that across all clusters the cluster assignment strength is high. Cluster utilization is slightly lower on the real-world datasets Mutagenicity and Reddit-Binary, reading 0.68 and 0.51, respectively. This can be attributed to the tasks being more difficult, model accuracy lower and the number of clusters potentially being set to high.  

\begin{table*}[ht]
\caption{Cluster utilization score of the Subgraph Concept Network (SCN) and a Concept Graph Network (CGN) using DiffPool \citep{diffpool}, measuring the average conditional cluster assignment strength per cluster to identify whether all clusters are used.}
\centering
\resizebox{0.8\textwidth}{!}{
\begin{tabular}{lll|ll}
\hline
 & \multicolumn{4}{c}{\textbf{\begin{tabular}[c]{@{}c@{}}Cluster Utilization\end{tabular}}} \\
 & \multicolumn{2}{c}{\textbf{SCN}} & \multicolumn{2}{c}{\textbf{CGN w/ DiffPool}} \\
 & \multicolumn{1}{c}{\textbf{Train}} & \multicolumn{1}{c}{\textbf{Test}} & \multicolumn{1}{|c}{\textbf{Train}} & \multicolumn{1}{c}{\textbf{Test}}     \\ 
 \hline
\textbf{Grid} & \textbf{0.97 (0.95, 0.99)} & \textbf{0.97 (0.95, 0.99)} & 0.50 (0.50, 0.50) & 0.50 (0.50, 0.50) \\
\textbf{Grid-House} & \textbf{0.91 (0.87, 0.95)} & \textbf{0.91 (0.87, 0.95)} & 0.25 (0.25, 0.25) & 0.25 (0.25, 0.25) \\
\textbf{STARS} & \textbf{0.99 (0.98, 0.99)} & \textbf{0.99 (0.98, 0.99)} & 0.50 (0.50, 0.50) & 0.50 (0.50, 0.50) \\
\textbf{House-Colour} & \textbf{0.90 (0.88, 0.92)} & \textbf{0.89 (0.87, 0.92)} & 0.25 (0.25, 0.25) & 0.25 (0.25, 0.25) \\
\hline
\textbf{Mutagenicity} & \textbf{0.68 (0.67, 0.69)} & \textbf{0.68 (0.67, 0.69)} & 0.23 (0.20, 0.27) & 0.19 (0.10, 0.29) \\
\textbf{Reddit-Binary} & \textbf{0.51 (0.45, 0.56)} & \textbf{0.51 (0.46, 0.56)} & 0.10 (0.10, 0.10) & 0.10 (0.10, 0.10) \\
\hline
\end{tabular}%
}

\label{fig:conditional_subgraph_assignment_strength}
\end{table*}

\subsection{Visualising subgraph-level concepts in subgraph concept networks}
\label{qualitive}

\begin{figure}[h]
\centerline{\includegraphics[width=0.9\textwidth]{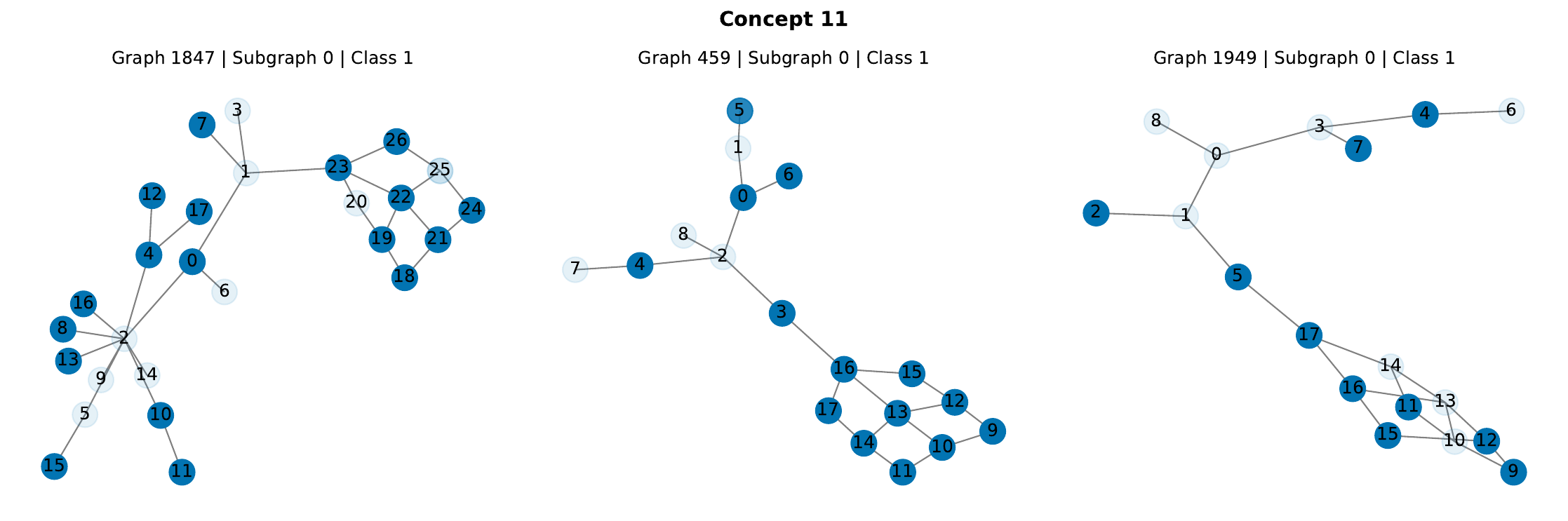}}
\caption{Subgraph concept discovered for the Grid dataset, showing that the grid structure is identified as a common subgraph across graphs.}
\label{subgraph_concept_grid}
\end{figure}

In comparison to the GCN-based baselines, the SCN allows to discover subgraph-level concepts. We focus on visualising and analysing subgraph-level concepts, as these are one of the novelties of the SCN. Recall, that node-level concepts can be visualised by first binarizing the concept encoding vector of the nodes, then computing the centroid of the nodes assigned to the same binarized concept and lastly visualising the $p$-hop neighbourhood of the nodes closest to the cluster centroid. We can also visualise subgraph-level concepts in the same manner with a slight modification. Instead of visualising the $p$-hop neighbourhood, we propose to visualise the full graph with the nodes coloured based on their assignment to the specific subgraph. Figure \ref{subgraph_concept_grid} is an example of a subgraph-level concept discovered for the Grid datasets. We can see that the grid structure is successfully recovered in Figure \ref{subgraph_concept_grid}.

We refer the reader to Appendix \ref{node_concept_vis} for more concept visualisations.

\begin{figure}[h]
    \centering
    \begin{minipage}[t]{0.22\textwidth}
        \vspace{0pt}
        \centering
        \subfloat[Cluster Activation]{%
            \includegraphics[width=\textwidth]{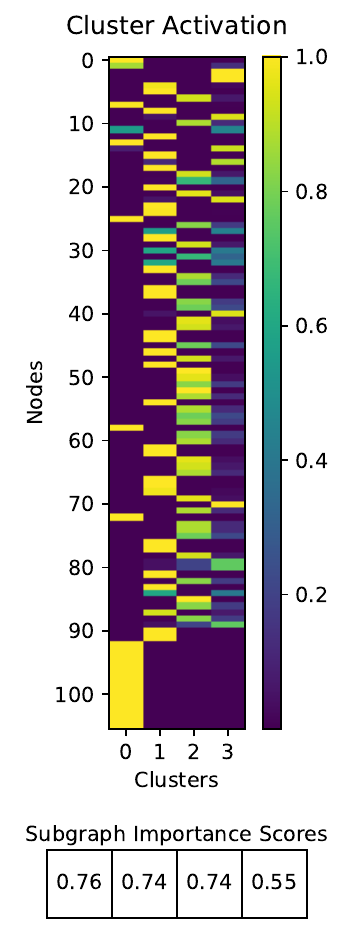}
            \label{fig:grid_house_a}
        }
    \end{minipage}%
    \hspace{0.01\textwidth}
    \begin{minipage}[t]{0.33\textwidth}
        \vspace{0pt}
        \centering
        \subfloat[Subgraph 1]{%
            \includegraphics[width=\textwidth]{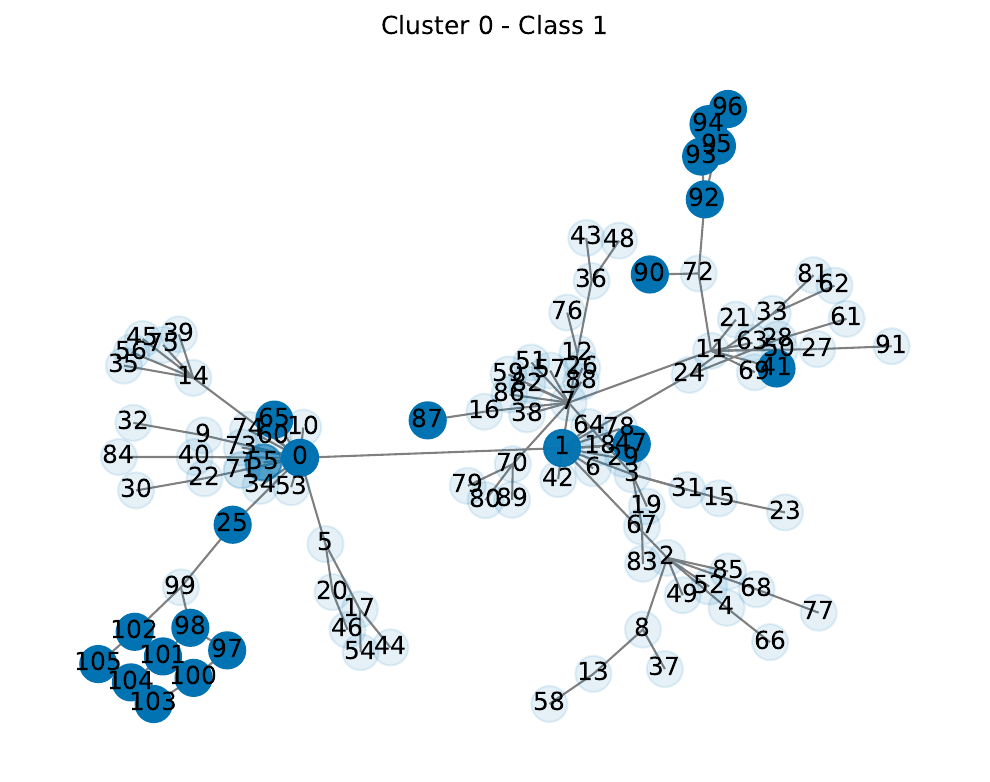}
            \label{fig:grid_house_b}
        }

        \vspace{0.1em}

        \subfloat[Subgraph 2]{%
            \includegraphics[width=\textwidth]{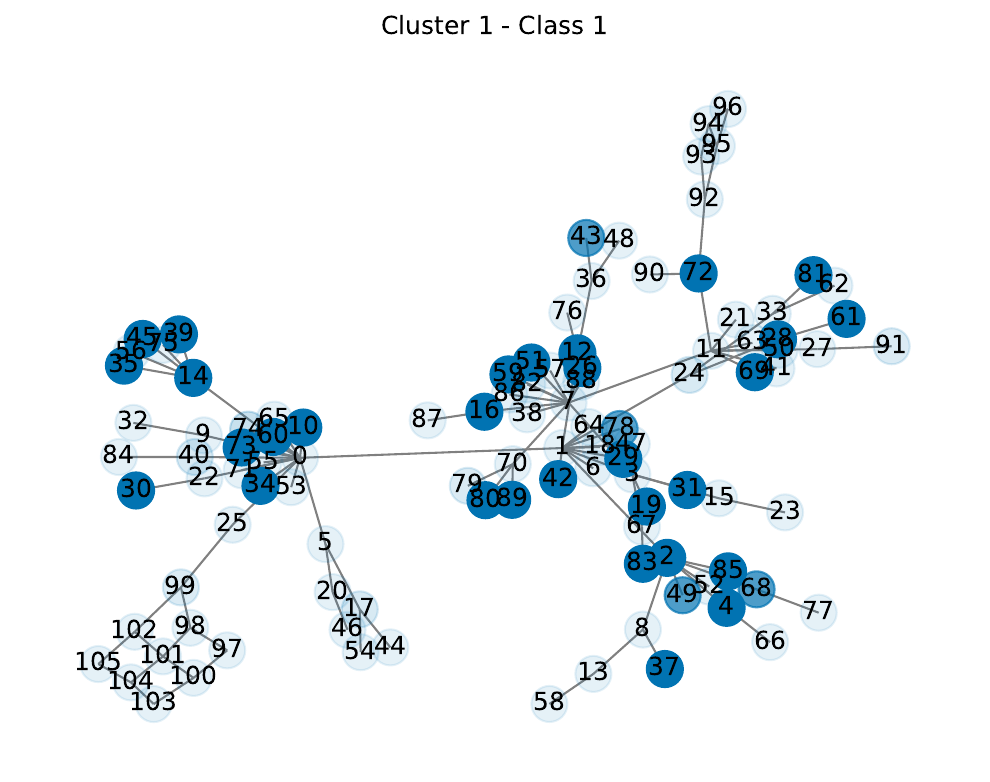}
            \label{fig:grid_house_c}
        }
    \end{minipage}%
    \hspace{0.01\textwidth}
    \begin{minipage}[t]{0.33\textwidth}
        \vspace{0pt}
        \centering
        \subfloat[Subgraph 3]{%
            \includegraphics[width=\textwidth]{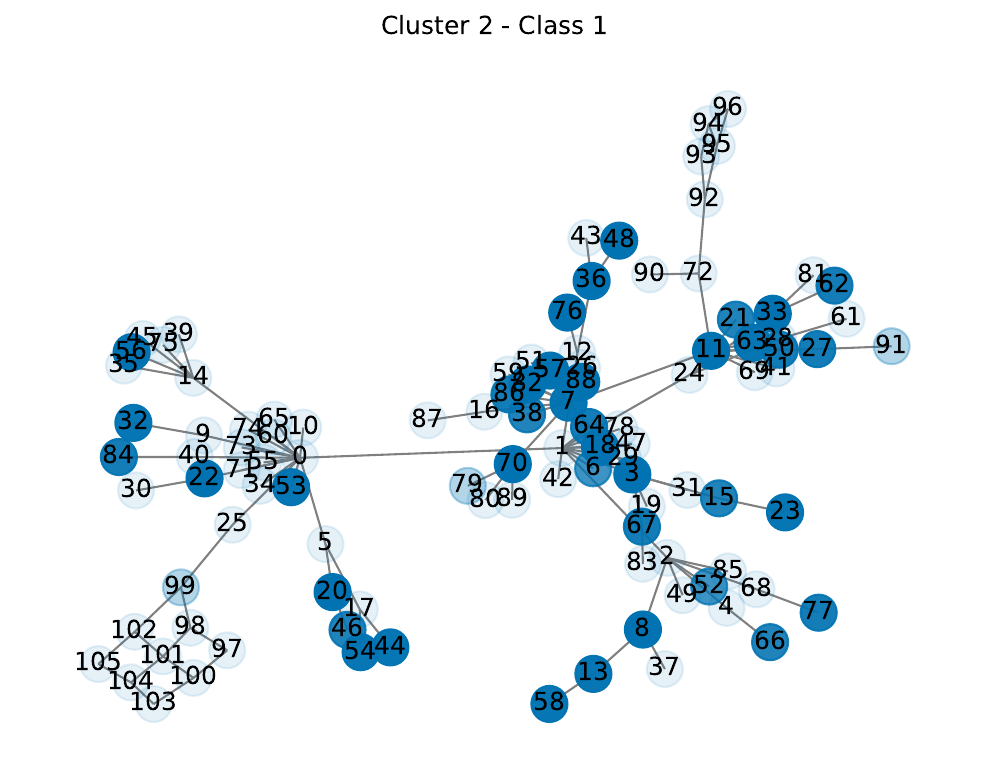}
            \label{fig:grid_house_d}
        }

        \vspace{0.1em}

        \subfloat[Subgraph 4]{%
            \includegraphics[width=\textwidth]{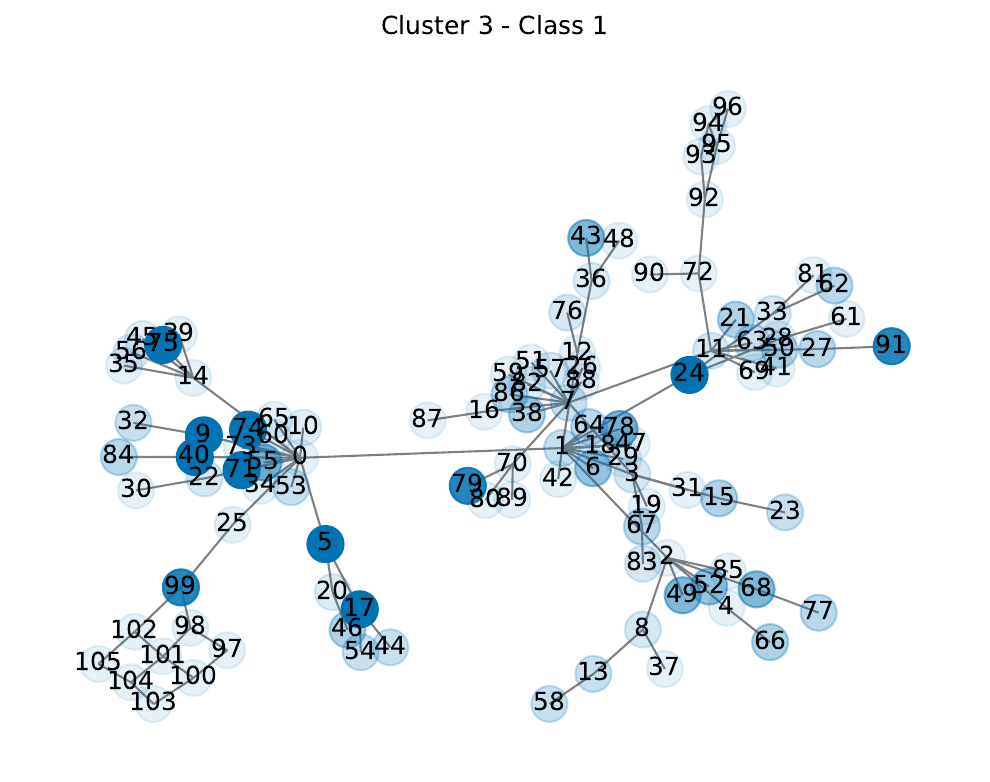}
            \label{fig:grid_house_e}
        }
    \end{minipage}
    \caption{Instance-level explanation for the Grid-House dataset. We can see the grid and house structure are part of Subgraph 1 and have the highest subgraph importance score. The base graph is split across subgraphs 2-4 with lower but still high subgraph importance scores.}
    \label{fig:grid_house_instance}
\end{figure}

While the aforementioned concept visualizations are at the global level, we can also obtain more detailed instance-level explanations. We can visualize the subgraphs extracted for a given input by plotting the subgraphs, the node assignment strengths and subgraph importance scores. For example, Figures \ref{fig:grid_house_instance}, \ref{fig:house_color_instance} and \ref{fig:reddit_instance} show examples for the Grid-House, House-Colour and Reddit-Binary datasets. In Figure \ref{fig:grid_house_instance} we can see that the SCN identifies the motifs encoded in the Grid-House dataset, namely the grid and house structure, in subgraph 1 (Figure \ref{fig:grid_house_b}) and assigns this subgraph the highest subgraph importance score at 0.76. The remaining subgraphs identified cover the base graph, with subgraph importance scores ranging from 0.55 to 0.74. The subgraphs are formed from mostly connected nodes, but not always. This indicates that the number of clusters was potentially set to low, so the salient motifs are grouped in one subgraph despite not being connected. However, it can also indicate a need for heavier regularization on subgraph connectivity, as the model may collapse to simply identifying the nodes relevant for prediction as one subgraph without caring about node connectivity. Moreover, we note that we do not restrict that a higher value indicates importance, rather than a value close to 0. We refer the reader to Appendix \ref{node_concept_vis} and \ref{qualitative_gip} for further analysis of concept visualisations and a qualitative comparison to the GIP method \citep{wang2024unveiling}.

\section{Discussion}
\label{discussion}

The SCN is the first approach to use subgraph-level concepts for more interpretable graph classification. Furthermore, it is the first approach to define the existence of different concept-levels, namely node, subgraph and graph-level concepts. The core limitation of the SCN is that it requires to define the number of clusters used for subgraph extraction. If the number of clusters is set too low, subgraphs may be grouped together, resulting in a larger subgraph that may obscure which structure is the most important. Defining the number of subgraphs is a common limitation of existing interpretable GNNs, such as prototype-based GNNs. However, in comparison to other methods, the SCN is not restricted across the subgraph concept space, unlike prototype-based models, which do not have the notion of concepts. The second main limitation of the SCN is that it assumes a form of ordering in the subgraphs extracted, because it uses a transpose of the importance scores computed from the pooled subgraph concepts as input for the readout layer. This requires the additional consistency loss term $\mathcal{L}_{\text{consistency}}$. Finally, a minor limitation of the SCN is that it relies on the same visualisation technique as GCExplainer \citep{magister2021gcexplainer} and CDM, simply visualising the concept as the $p$-hop neighbourhood. However, SCN offers new visualizations of the salient subgraphs for prediction. We note, that as GNNs become more powerful and interpretable, we must stay vigilant to avoid potential risks or societal harm.

\section{Conclusion}
\label{conclusion}

We demonstrate that SCNs provide an increased insight into the model's reasoning process by introducing subgraph and graph-level concepts. To the best of our knowledge, our approach is the first to allow interpreting a GNN via different levels of concepts, namely node, subgraph and graph-level concepts, while maintaining task performance. Our experiments show that the concepts are meaningful and interpretable and can be used to explain different slices of the model. Furthermore, the subgraph-level concepts and subgraph importance scores allow to visualise subgraph level concepts in a different and more meaningful manner than simply the $p$-hop neighbourhood.

\bibliographystyle{unsrtnat}
\bibliography{references}

@article{tishby2000information,
  title={The information bottleneck method},
  author={Tishby, Naftali and Pereira, Fernando C and Bialek, William},
  journal={arXiv preprint physics/0004057},
  year={2000}
}

@article{magister2021gcexplainer,
  title={GCExplainer: Human-in-the-Loop Concept-based Explanations for Graph Neural Networks},
  author={Magister, Lucie Charlotte and Kazhdan, Dmitry and Singh, Vikash and Li{\`o}, Pietro},
  journal={arXiv preprint arXiv:2107.11889},
  year={2021}
}

@article{ying2019gnnexplainer,
  title={Gnnexplainer: Generating explanations for graph neural networks},
  author={Ying, Zhitao and Bourgeois, Dylan and You, Jiaxuan and Zitnik, Marinka and Leskovec, Jure},
  journal={Advances in neural information processing systems},
  volume={32},
  year={2019}
}

@article{morris2020tudataset,
  title={Tudataset: A collection of benchmark datasets for learning with graphs},
  author={Morris, Christopher and Kriege, Nils M and Bause, Franka and Kersting, Kristian and Mutzel, Petra and Neumann, Marion},
  journal={arXiv preprint arXiv:2007.08663},
  year={2020}
}

@article{scarselli2008graph,
  title={The graph neural network model},
  author={Scarselli, Franco and Gori, Marco and Tsoi, Ah Chung and Hagenbuchner, Markus and Monfardini, Gabriele},
  journal={IEEE transactions on neural networks},
  volume={20},
  number={1},
  pages={61--80},
  year={2008},
  publisher={IEEE}
}

@book{breiman1984classification,
  title={Classification and Regression Trees},
  author={Breiman, L and Friedman, JH and Olshen, R and Stone, CJ},
  year={1984},
  publisher={Wadsworth}
}

@inproceedings{zhang2022protgnn,
  title={Protgnn: Towards self-explaining graph neural networks},
  author={Zhang, Zaixi and Liu, Qi and Wang, Hao and Lu, Chengqiang and Lee, Cheekong},
  booktitle={Proceedings of the AAAI Conference on Artificial Intelligence},
  volume={36},
  pages={9127--9135},
  year={2022}
}

@inproceedings{yeh2020completeness,
author = {Yeh, Chih-Kuan and Kim, Been and Arık, Sercan {\"{O}} and Li, Chun-Liang and Pfister, Tomas and Ravikumar, Pradeep},
booktitle = {Advances in Neural Information Processing Systems 33 (NeurIPS 2020)},
pages = {20554--20565},
title = {{On Completeness-aware Concept-Based Explanations in Deep Neural Networks}},
url = {https://papers.nips.cc/paper/2020/hash/ecb287ff763c169694f682af52c1f309-Abstract.html},
volume = {33},
year = {2020}
}

@inproceedings{azzolin2023globalexplainabilitygnnslogic,
title={Global Explainability of GNNs via Logic Combination of Learned Concepts}, 
author={Steve Azzolin and Antonio Longa and Pietro Barbiero and Pietro Liò and Andrea Passerini},
booktitle={International Conference on Learning Representations},
year={2023},
organization={PMLR}
}

@inproceedings{diffpool, author = {Ying, Rex and You, Jiaxuan and Morris, Christopher and Ren, Xiang and Hamilton, William L. and Leskovec, Jure}, title = {Hierarchical graph representation learning with differentiable pooling}, year = {2018}, publisher = {Curran Associates Inc.}, address = {Red Hook, NY, USA}, booktitle = {Proceedings of the 32nd International Conference on Neural Information Processing Systems}, pages = {4805–4815}, numpages = {11}, location = {Montr\'{e}al, Canada}, series = {NIPS'18} }

@inproceedings{
Yuan2020StructPool,
title={StructPool: Structured Graph Pooling via Conditional Random Fields},
author={Hao Yuan and Shuiwang Ji},
booktitle={International Conference on Learning Representations},
year={2020},
url={https://openreview.net/forum?id=BJxg_hVtwH}
}

@article{2023everybodyneedslittlehelp,
  title={Everybody Needs a Little HELP: Explaining Graphs via Hierarchical Concepts},
  author={Jurss, Jonas and Magister, Lucie Charlotte and Barbiero, Pietro and Li{\`o}, Pietro and Simidjievski, Nikola},
  journal={arXiv preprint arXiv:2311.15112},
  year={2023}
}

@article{ragno2024,
  author={Ragno, Alessio and Rosa, Biagio La and Capobianco, Roberto},
  journal={IEEE Transactions on Artificial Intelligence}, 
  title={Prototype-Based Interpretable Graph Neural Networks}, 
  year={2024},
  volume={5},
  number={4},
  pages={1486-1495},
 }

@article{chen2019,
  title={This looks like that: deep learning for interpretable image recognition},
  author={Chen, Chaofan and Li, Oscar and Tao, Daniel and Barnett, Alina and Rudin, Cynthia and Su, Jonathan K},
  journal={Advances in neural information processing systems},
  volume={32},
  year={2019}
}

@inproceedings{wang2021tesnet,
  title={Interpretable image recognition by constructing transparent embedding space},
  author={Wang, Jiaqi and Liu, Huafeng and Wang, Xinyue and Jing, Liping},
  booktitle={Proceedings of the IEEE/CVF international conference on computer vision},
  pages={895--904},
  year={2021}
}

@inproceedings{dcr2023, author = {Barbiero, Pietro and Ciravegna, Gabriele and Giannini, Francesco and Zarlenga, Mateo Espinosa and Magister, Lucie Charlotte and Tonda, Alberto and Li\'{o}, Pietro and Precioso, Frederic and Jamnik, Mateja and Marra, Giuseppe}, title = {Interpretable neural-symbolic concept reasoning}, year = {2023}, publisher = {JMLR.org}, booktitle = {Proceedings of the 40th International Conference on Machine Learning}, articleno = {76}, numpages = {25}, location = {Honolulu, Hawaii, USA}, series = {ICML'23} }

@article{entropy,
  author={Shannon, C. E.},
  journal={The Bell System Technical Journal}, 
  title={A mathematical theory of communication}, 
  year={1948},
  volume={27},
  number={3},
  pages={379-423},
  doi={10.1002/j.1538-7305.1948.tb01338.x}}

@article{longa2022explaining,
  title={Explaining the explainers in graph neural networks: a comparative study},
  author={Longa, Antonio and Azzolin, Steve and Santin, Gabriele and Cencetti, Giulia and Lio, Pietro and Lepri, Bruno and Passerini, Andrea},
  journal={ACM Computing Surveys},
  volume={57},
  number={5},
  pages={1--37},
  year={2025},
  publisher={ACM New York, NY}
}

@article{Erdos1984OnTE,
  title={On the evolution of random graphs},
  author={Paul L. Erdos and Alfr{\'e}d R{\'e}nyi},
  journal={Transactions of the American Mathematical Society},
  year={1984},
  volume={286},
  pages={257-257},
  url={https://api.semanticscholar.org/CorpusID:6829589}
}

@article{corso2024graph,
  title={Graph neural networks},
  author={Corso, Gabriele and Stark, Hannes and Jegelka, Stefanie and Jaakkola, Tommi and Barzilay, Regina},
  journal={Nature Reviews Methods Primers},
  volume={4},
  number={1},
  pages={17},
  year={2024},
  publisher={Nature Publishing Group UK London}
}

@article{ji2025comprehensive,
  title={A comprehensive survey on self-interpretable neural networks},
  author={Ji, Yang and Sun, Ying and Zhang, Yuting and Wang, Zhigaoyuan and Zhuang, Yuanxin and Gong, Zheng and Shen, Dazhong and Qin, Chuan and Zhu, Hengshu and Xiong, Hui},
  journal={Proceedings of the IEEE},
  year={2025},
  publisher={IEEE}
}

@inproceedings{lee2025pre,
  title={Pre-training graph neural networks on molecules by using subgraph-conditioned graph information bottleneck},
  author={Lee, O-Joun and others},
  booktitle={Proceedings of the AAAI Conference on Artificial Intelligence},
  volume={39},
  pages={17204--17213},
  year={2025}
}

@article{yu2021recognizing,
  title={Recognizing predictive substructures with subgraph information bottleneck},
  author={Yu, Junchi and Xu, Tingyang and Rong, Yu and Bian, Yatao and Huang, Junzhou and He, Ran},
  journal={IEEE transactions on pattern analysis and machine intelligence},
  volume={46},
  number={3},
  pages={1650--1663},
  year={2021},
  publisher={IEEE}
}

@inproceedings{yu2022improving,
  title={Improving subgraph recognition with variational graph information bottleneck},
  author={Yu, Junchi and Cao, Jie and He, Ran},
  booktitle={Proceedings of the IEEE/CVF conference on computer vision and pattern recognition},
  pages={19396--19405},
  year={2022}
}

@inproceedings{wang2024unveiling,
  title={Unveiling global interactive patterns across graphs: Towards interpretable graph neural networks},
  author={Wang, Yuwen and Liu, Shunyu and Zheng, Tongya and Chen, Kaixuan and Song, Mingli},
  booktitle={Proceedings of the 30th ACM SIGKDD conference on knowledge discovery and data mining},
  pages={3277--3288},
  year={2024}
}

@article{yang2025gnns,
  title={From GNNs to trees: Multi-granular interpretability for graph neural networks},
  author={Yang, Jie and Wang, Yuwen and Chen, Kaixuan and Zheng, Tongya and Zhou, Yihe and Xiao, Zhenbang and Cao, Ji and Song, Mingli and Liu, Shunyu},
  journal={arXiv preprint arXiv:2505.00364},
  year={2025}
}

@article{wu2020graph,
  title={Graph information bottleneck},
  author={Wu, Tailin and Ren, Hongyu and Li, Pan and Leskovec, Jure},
  journal={Advances in Neural Information Processing Systems},
  volume={33},
  pages={20437--20448},
  year={2020}
}

@article{graphxai_survey,
	author = {Nandan, Mauparna and Mitra, Soma and De, Debashis},
	isbn = {1433-3058},
	journal = {Neural Computing and Applications},
	number = {17},
	pages = {10949--11000},
	title = {GraphXAI: a survey of graph neural networks (GNNs) for explainable AI (XAI)},
	url = {https://doi.org/10.1007/s00521-025-11054-3},
	volume = {37},
	year = {2025},
}

@article{seo2023interpretable,
  title={Interpretable prototype-based graph information bottleneck},
  author={Seo, Sangwoo and Kim, Sungwon and Park, Chanyoung},
  journal={Advances in Neural Information Processing Systems},
  volume={36},
  pages={76737--76748},
  year={2023}
}

@inproceedings{magister2023concept,
  title={Concept distillation in graph neural networks},
  author={Magister, Lucie Charlotte and Barbiero, Pietro and Kazhdan, Dmitry and Siciliano, Federico and Ciravegna, Gabriele and Silvestri, Fabrizio and Jamnik, Mateja and Li{\`o}, Pietro},
  booktitle={World Conference on Explainable Artificial Intelligence},
  pages={233--255},
  year={2023},
  organization={Springer}
}

\newpage

\appendix

\section{Dataset statistics}
\label{dataset_stats}

We summarise important dataset statistics in Table \ref{dataset_sizes}.

\begin{table}[h]
\caption{An overview of key markers of the datasets.}
\centering
\resizebox{0.8\textwidth}{!}{%
\begin{tabular}{lcccc}
\hline
\multicolumn{1}{c}{\textbf{Dataset}} &
  \multicolumn{1}{c}{\textbf{\begin{tabular}[c]{@{}c@{}}Number of\\ Graphs\end{tabular}}} &
  \multicolumn{1}{c}{\textbf{\begin{tabular}[c]{@{}c@{}}Average\\ Graph Size\end{tabular}}} &
  \multicolumn{1}{c}{\textbf{\begin{tabular}[c]{@{}c@{}}Number of\\ Features\end{tabular}}} &
  \multicolumn{1}{c}{\textbf{\begin{tabular}[c]{@{}c@{}}Number of\\ Classes\end{tabular}}} \\ \hline
\textbf{Grid}    & 2000   &   22.17  & 1  & 2 \\
\textbf{Grid-House}   & 1000    &  122.82 & 1  & 2 \\
\textbf{STARS}        & 1500    & 63.92 & 1  & 3 \\
\textbf{House-Colour}    & 1000    & 46.95  & 3  & 2 \\
\textbf{Mutagenicity}   & 4337 & 30.32  & 14 & 2 \\
\textbf{Reddit-Binary}  & 2000 & 429.63 & 1  & 2 \\ \hline
\end{tabular}%
}
\label{dataset_sizes}
\end{table}

\section{Model training}
\label{model_stats}

Table \ref{table:models} summarises the model parametrisation used across the experiments. We keep the number of initial graph embedding layers consistent across the models to provide them with the same representational power and keep all training hyperparameters fixed. Our hyperparameter choices were informed by the study of \citet{longa2022explaining} on the performance of GNNs on the synthetic datasets and CDM \citep{magister2023concept}. We train each model across five different seeds to allow reporting confidence intervals. We fix the seeds for repoducibility: 42, 76, 58, 92, 19,  

\begin{table*}[h]
\caption{A summary of the model architecture and training hyperparameters for each dataset, following recommendations from \citet{longa2022explaining}. These parameters are kept the same between the Subgraph Concept Network and Concept Graph Network variants.}
\centering
\resizebox{\textwidth}{!}{
\begin{tabular}{lcccccccc}
\hline
\textbf{Dataset} &
  \textbf{\begin{tabular}[c]{@{}c@{}}Number of \\ Graph Conv.\end{tabular}} &
  \textbf{\begin{tabular}[c]{@{}c@{}}Number of \\ Hidden Units\end{tabular}} &
  \textbf{\begin{tabular}[c]{@{}c@{}}Node Concept \\ Encoding Size\end{tabular}} &
  \textbf{\begin{tabular}[c]{@{}c@{}}Number of \\ Subgraphs \end{tabular}} &
  \textbf{\begin{tabular}[c]{@{}c@{}}Subgraph Concept \\ Encoding Size\end{tabular}} &
  \textbf{\begin{tabular}[c]{@{}c@{}}Learning \\ Rate\end{tabular}} &
  \textbf{\begin{tabular}[c]{@{}c@{}}Batch \\ Size\end{tabular}} &
  \textbf{\begin{tabular}[c]{@{}c@{}}Number of \\ Epochs\end{tabular}} \\ \hline
\textbf{Grid}    & 5   &   20  & 10  & 2 & 8  & 0.001 & 16 & 300 \\
\textbf{Grid-House}   & 4    &  20 & 10  & 4 & 10 & 0.001 & 16 & 2000 \\
\textbf{STARS}        & 2    & 10 & 4  & 2 & 4 & 0.001 & 16 & 300 \\
\textbf{House-Colour}    & 2    & 10  & 10 & 4 & 8 & 0.001 & 16 & 300 \\
\textbf{Mutagenicity}   & 3 & 40 & 10 & 10 & 10 & 0.001 & 16 & 1000 \\
\textbf{Reddit-Binary}  & 3 & 32 & 10  & 10 & 10 & 0.001 & 16 & 1000 \\ \hline
\end{tabular}%
}

\label{table:models}
\end{table*}

\section{Node-level and graph-level concept completeness}
\label{node_level_concept_completeness}

\paragraph{Subgraph concept networks discover a comprehensive set of node-level concepts across subgraphs} Recall, that our architecture extracts node-level concepts at two levels, the full input graph and the subgraphs. First, we extract node-level concepts across the graph, which we then use in soft clustering. We later also extract node-level concepts after we have embedded the individual subgraphs. Our experiments show that the SCN discovers a comprehensive but not complete set of node-level concepts across the initial graph and subgraphs. In reference to Table \ref{fig:completeness_accuracy}, the SCN discovers the most complete set of node-level concepts at the subgraph level on the Grid, Grid-House, STARS, House-Colour and Reddit-Binary datasets out of the three models compared. We observe that the set of concepts discovered in the subgraph space is more complete and representative of the task than in the graph space. This may be attributed to more graph convolutional processing being carried out and less nodes contributing to message passing. While the SCN has the highest concept completeness considering concepts extracted in the subgraph space, concept completeness falls short of the predictive accuracy of the model. For example, the SCN only achieves a node-level concept completeness of 76.02\%, compared to a model accuracy of 99.40\%. This can be attributed to an individual node not being predictive of the final target in graph classification, as it is in node classification. Nevertheless, the SCN provides the most complete node-level concept representation out of the models compared. Note, that we cannot compute concept completeness for the GIB method, because it is not a concept-based method.

\begin{table*}[ht]
\caption{Node-level concept completeness for the Subgraph Concept Network (SCN) and two equivalent Concept Graph Networks (CGN) using mean pooling and DiffPool, respectively, as well as the GIP method \citep{wang2024unveiling}. Note that we can compute concept completeness for node-level concepts extracted at the graph and subgraph leve in the SCN.}
\centering
\resizebox{1\textwidth}{!}{
\begin{tabular}{lllll}
\hline     & \multicolumn{4}{c}{\textbf{\begin{tabular}[c]{@{}c@{}}Node-level Concept Completeness (\%)\end{tabular}}} \\
                            &\multicolumn{1}{c}{\begin{tabular}[c]{@{}c@{}}\textbf{SCN in}\\\textbf{graph space}\end{tabular}}
                            & \multicolumn{1}{c}{\begin{tabular}[c]{@{}c@{}}\textbf{SCN in}\\\textbf{subgraph space}\end{tabular}}
                            & \multicolumn{1}{c}{\begin{tabular}[c]{@{}c@{}}\textbf{CGN}\\\textbf{w/ mean pool}\end{tabular}}
                            & \multicolumn{1}{c}{\begin{tabular}[c]{@{}c@{}}\textbf{CGN}\\\textbf{w/ DiffPool}\end{tabular}} \\ \hline
\textbf{Grid} & 54.06 (43.86, 64.27) & \textbf{76.02 (62.89, 89.15)} & 70.44 (69.82, 71.06) & 67.06 (60.66, 73.47)                  \\
\textbf{Grid-House} & 51.71 (51.68, 51.75) & \textbf{54.95 (51.78, 58.13)} & 51.63 (51.47, 51.79) & 50.92 (50.34, 51.50) \\
\textbf{STARS} & 34.96 (32.54, 37.37) & \textbf{61.54 (51.47, 71.62)} & 47.79 (45.79, 49.80) & 47.13 (45.22, 49.04)                \\
\textbf{House-Colour} & 53.84 (45.17, 62.52) & \textbf{78.01 (62.15, 93.88)} & 55.25 (50.85, 59.64) & 52.96 (49.51, 56.41)              \\
\hline
\textbf{Mutagenicity}  & 60.39 (55.53, 65.25) & 59.39 (57.89, 60.89) & 63.55 (62.21, 64.88) & \textbf{59.46 (56.09, 62.83)}    \\
\textbf{Reddit-Binary} & 61.86 (57.68, 66.04) & \textbf{70.22 (63.50, 76.93)} & 69.71 (64.94, 74.48) & 60.83 (57.51, 64.14)                  \\
\hline
\end{tabular}%
}

\label{fig:completeness_accuracy}
\end{table*}

\paragraph{Subgraph concept networks discover a complete set graph-level concepts} In reference to Table \ref{fig:completeness_accuracy3}, our results show that SCNs discover a complete set of graph-level concepts. While the CGN using DiffPool \citep{diffpool} outperforms the SCN on four of the six datasets, the graph-level concept completeness of the SCN is very comparable. For example, on the real-world dataset Mutagenicity the SCN achieves a graph-level concept completeness score of 72.81\%, while the CGN with DiffPool achieves a score of 74.53\%. The slightly sub-par performance of the SCN can potentially be attributed to the model relying only on the subgraph importance scores for prediction, rather than the subgraph embeddings. This was a conscious design choice as mean pooling the subgraph embeddings would obscure the interpretability of the model, while concatenating the subgraph embeddings would quickly increase the hidden dimension of the readout layer. Moreover, the clustered importance scores provide a more meaningful representation for interpretation than the mean pooled coarsened subgraph embeddings of the CGN using DiffPool. It can therefore be said that the SCN comes with increased interpretability at the trade-off of slighlty lower graph-level concept completeness in comparison to the CGN.

We can compute graph-level concept completeness for the GIP method\citep{wang2024unveiling} via a slight extension. While the GIP method does not operate in the concept space, but rather learns prototypes, it produces similarity scores by matching the prototypes to subgraphs in the input graph. Since the graph-level concept embeddings for the SCN are derived from the subgraph importance scores, the similarity scores of the GIP method can be seen similarly. However, we note that the GIP architecture does not optimize for concepts like SCN. Nevertheless, if we treat the graph similarity scores as graph-level concepts, the GIP outperforms the SCN in respect to graph-level concept completeness. This may be attributed to prototypes being fixed learned representations from which the similarity scores are computed, while the SCN is less regulated.

\begin{table*}[ht]
\caption{Graph-level concept completeness for the Subgraph Concept Network (SCN) and a Concept Graph Network (CGN) using mean pool and DiffPool \citep{diffpool}. We perform post-processing to also extend the computation of the graph-level concept completeness to the GIP method \citep{wang2024unveiling}.}
\centering
\resizebox{1\textwidth}{!}{
\begin{tabular}{lllll}
\hline     & \multicolumn{4}{c}{\textbf{\begin{tabular}[c]{@{}c@{}}Graph-level Concept  Completeness (\%)\end{tabular}}} \\
                            & \multicolumn{1}{c}{\textbf{SCN}}          
                            & \multicolumn{1}{c}{\begin{tabular}[c]{@{}c@{}}\textbf{CGN}\\\textbf{w/ mean pool}\end{tabular}}
                            & \multicolumn{1}{c}{\begin{tabular}[c]{@{}c@{}}\textbf{CGN}\\\textbf{w/ DiffPool}\end{tabular}}
                            & \multicolumn{1}{c}{\begin{tabular}[c]{@{}c@{}}\textbf{GIP}\\\textbf{w/ similarity scores}\end{tabular}}\\ \hline
\textbf{Grid} & 93.55 (88.58, 98.52) & 84.00 (62.05, 100.00) & 84.15 (57.43, 100.00) & \textbf{99.80
(99.54, 100.00)}             \\
\textbf{Grid-House} & 69.80 (56.87, 82.73) & 58.50 (42.53, 74.47) & 50.00 (50.00, 50.00) & \textbf{99.30 (98.74, 99.86)} \\
\textbf{STARS} & 89.13 (70.21, 100.00) & 41.53 (25.20, 57.87) & \textbf{99.13 (98.66, 99.60)} & 97.73 (96.84, 98.63)                \\
\textbf{House-Colour} & 97.90 (96.18, 99.62) & 46.20 (42.81, 49.59) & 97.90 (96.98, 98.82) & \textbf{99.30 (98.59, 100.00)} \\
\hline
\textbf{Mutagenicity}  & 72.81 (65.22, 80.41) & 59.44 (57.64, 61.25) & \textbf{74.53 (68.76, 80.31)} & 72.70 (71.61, 73.79)            \\
\textbf{Reddit-Binary}  & 86.54 (80.46, 92.62) & 61.91 (59.36, 64.46) & \textbf{87.65 (85.44, 89.87)} & 84.01 (78.95, 89.07)                   \\
\hline
\end{tabular}%
}

\label{fig:completeness_accuracy3}
\end{table*}

\section{Further concept visualisations}
\label{node_concept_vis}

\paragraph{Subgraph-level concept visualisations}

\begin{figure}[h]
\centerline{\includegraphics[width=1\textwidth]{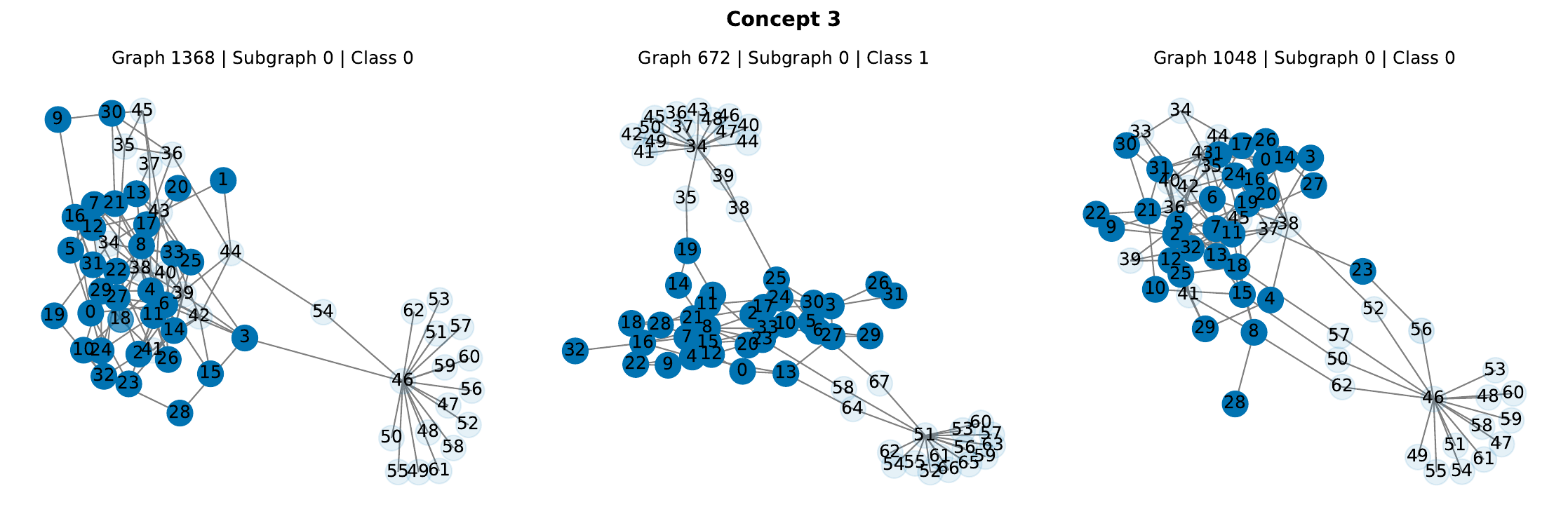}}
\caption{Subgraph concept discovered for the STARS dataset, showing that the model can separate the base graph from the star structure attached across classes.}
\label{subgraph_concept_stars}
\end{figure}

\begin{figure}[h]
    \centering
    \begin{minipage}[t]{0.23\textwidth}
        \vspace{0pt}
        \centering
        \subfloat[Cluster Activation]{%
            \includegraphics[width=\textwidth]{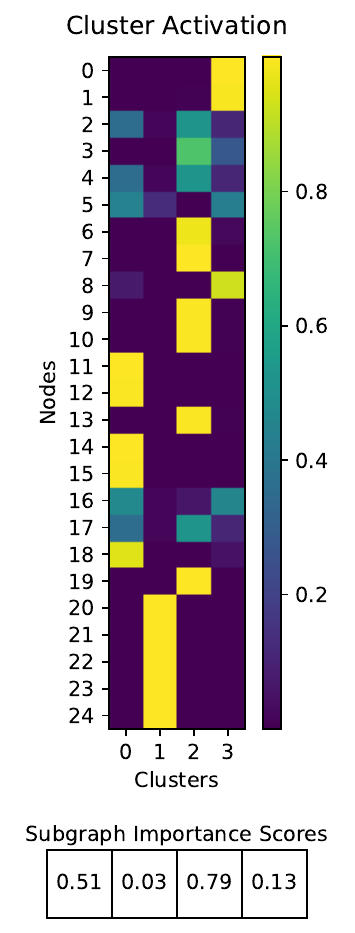}
            \label{fig:house_color_a}
        }
    \end{minipage}%
    \hspace{0.01\textwidth}
    \begin{minipage}[t]{0.33\textwidth}
        \vspace{0pt}
        \centering
        \subfloat[Subgraph 1]{%
            \includegraphics[width=\textwidth]{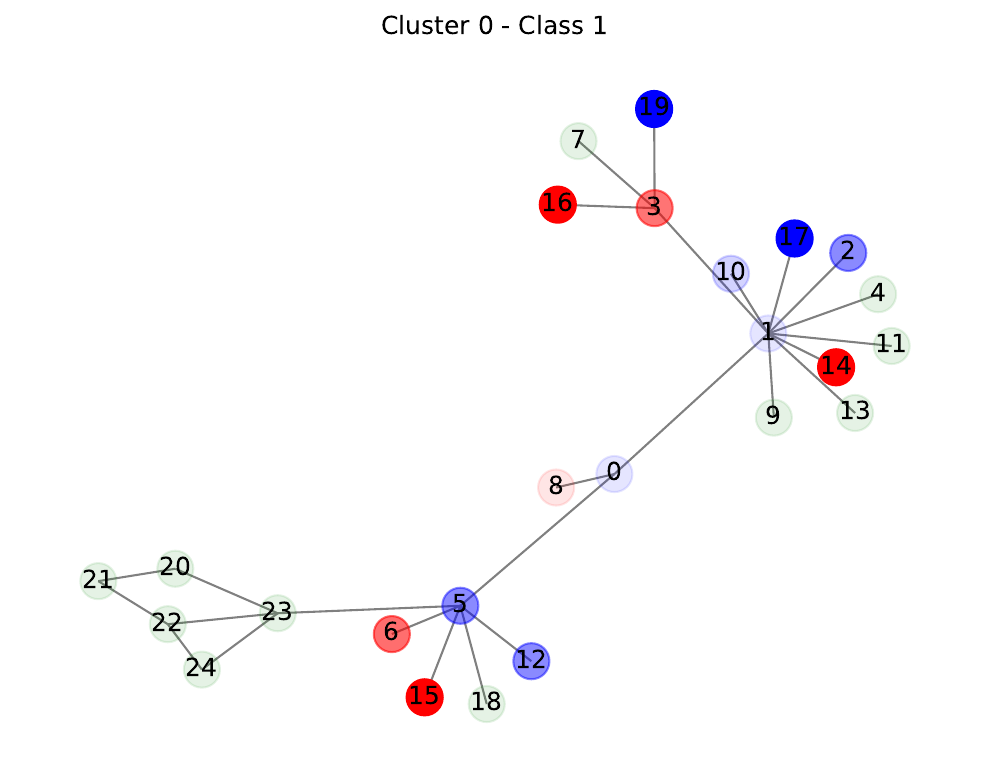}
            \label{fig:house_color_b}
        }

        \vspace{0.1em}

        \subfloat[Subgraph 2]{%
            \includegraphics[width=\textwidth]{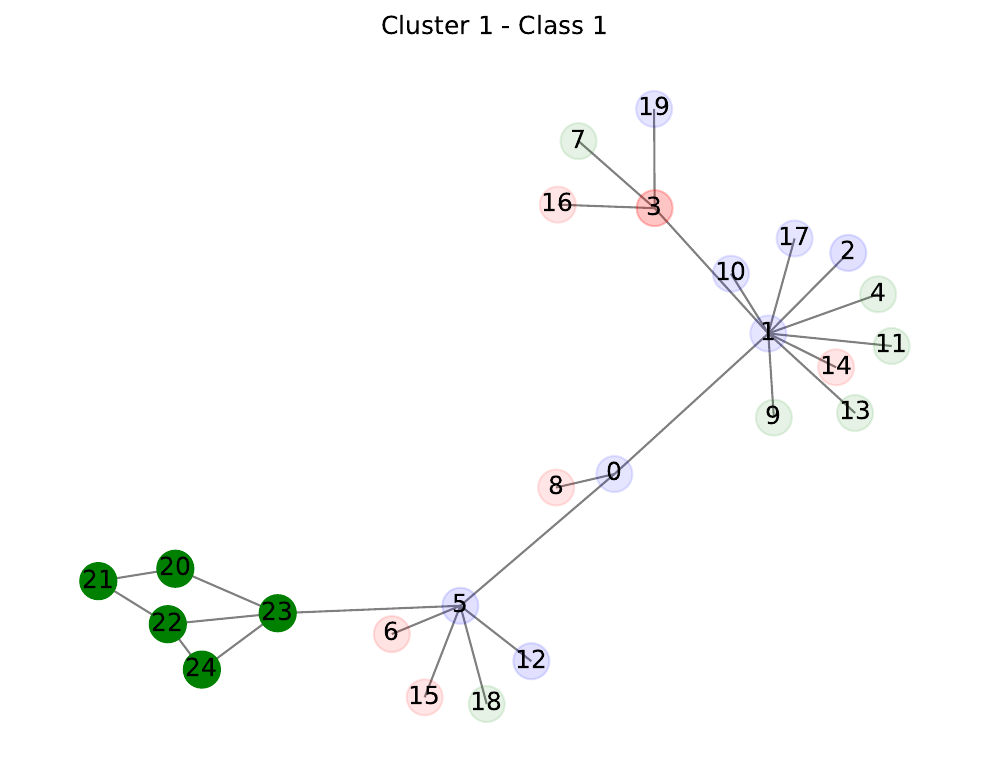}
            \label{fig:house_color_c}
        }
    \end{minipage}%
    \hspace{0.01\textwidth}
    \begin{minipage}[t]{0.33\textwidth}
        \vspace{0pt}
        \centering
        \subfloat[Subgraph 3]{%
            \includegraphics[width=\textwidth]{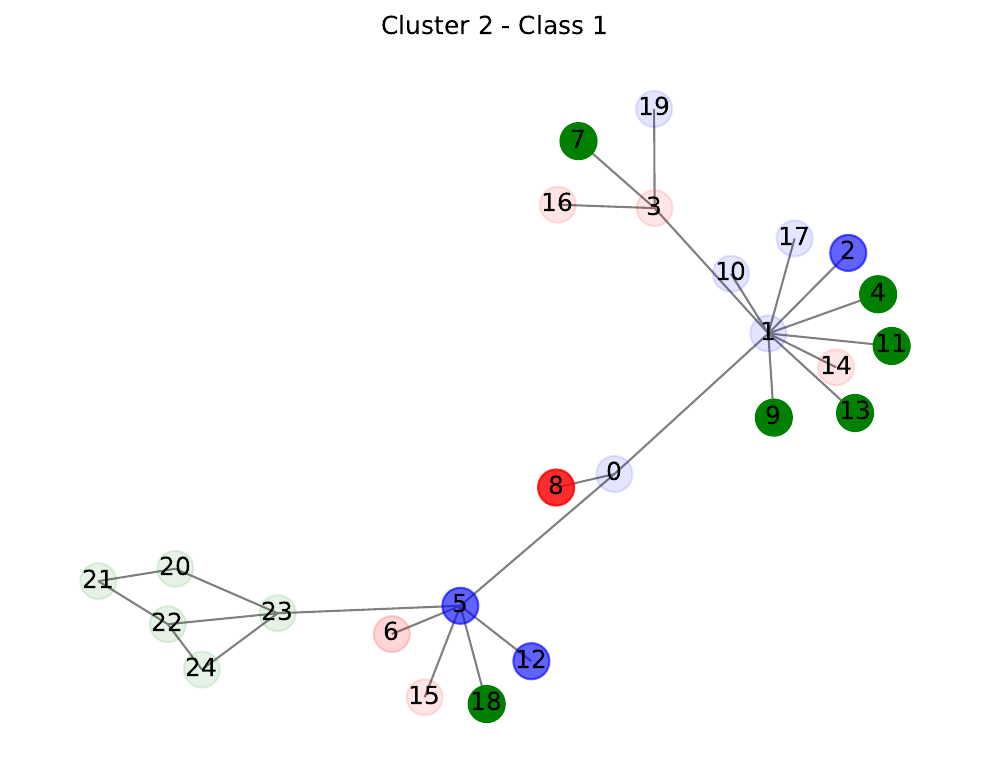}
            \label{fig:house_color_d}
        }

        \vspace{0.1em}

        \subfloat[Subgraph 4]{%
            \includegraphics[width=\textwidth]{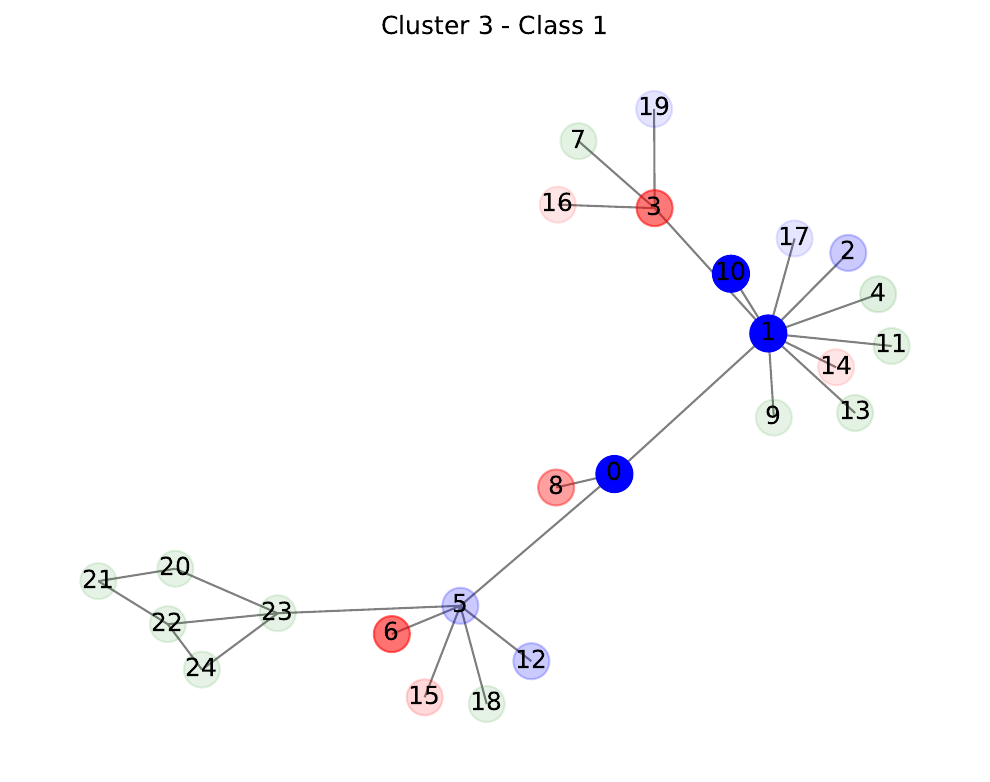}
            \label{fig:house_color_e}
        }
    \end{minipage}
    \caption{Instance-level explanation for the House-Colour dataset. We can see the green house structure is identified as Subgraph 2. Note that the subgraph importance score here is the lowest, however, we do not enforce the meaning of the subgraph importance score to be tied to 0 or 1, wherefore, the model may also interpret a value close to 0 as meaning the most important.}
    \label{fig:house_color_instance}
\end{figure}

\begin{figure}[h]
    \centering
    \begin{minipage}[t]{0.35\textwidth}
        \vspace{0pt}
        \centering
        \subfloat[Cluster Activation]{%
            \includegraphics[width=\textwidth]{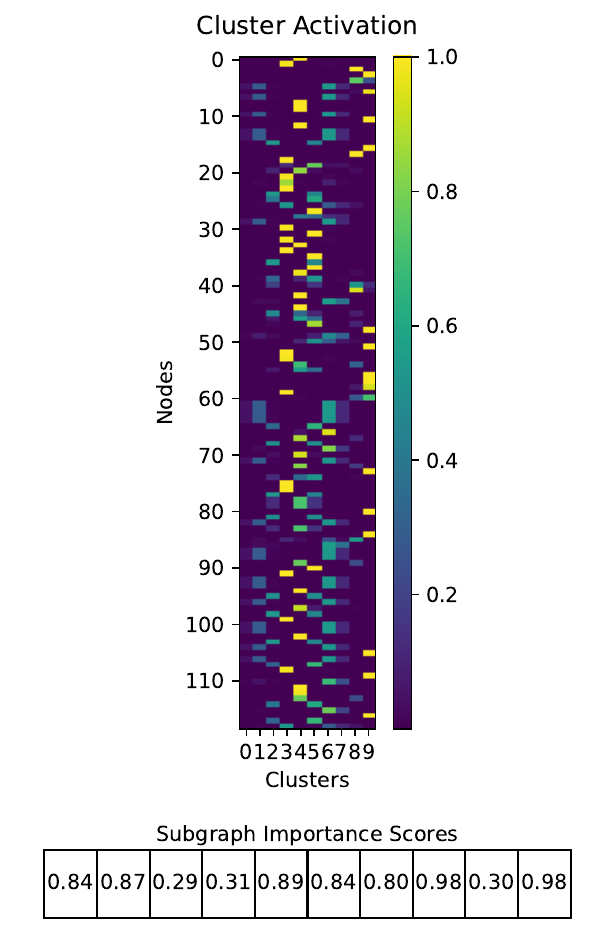}
            \label{fig:reddit_a}
        }
    \end{minipage}%
    \hspace{0.01\textwidth}
    \begin{minipage}[t]{0.3\textwidth}
        \vspace{0pt}
        \centering
        \subfloat[Subgraph 4]{%
            \includegraphics[width=\textwidth]{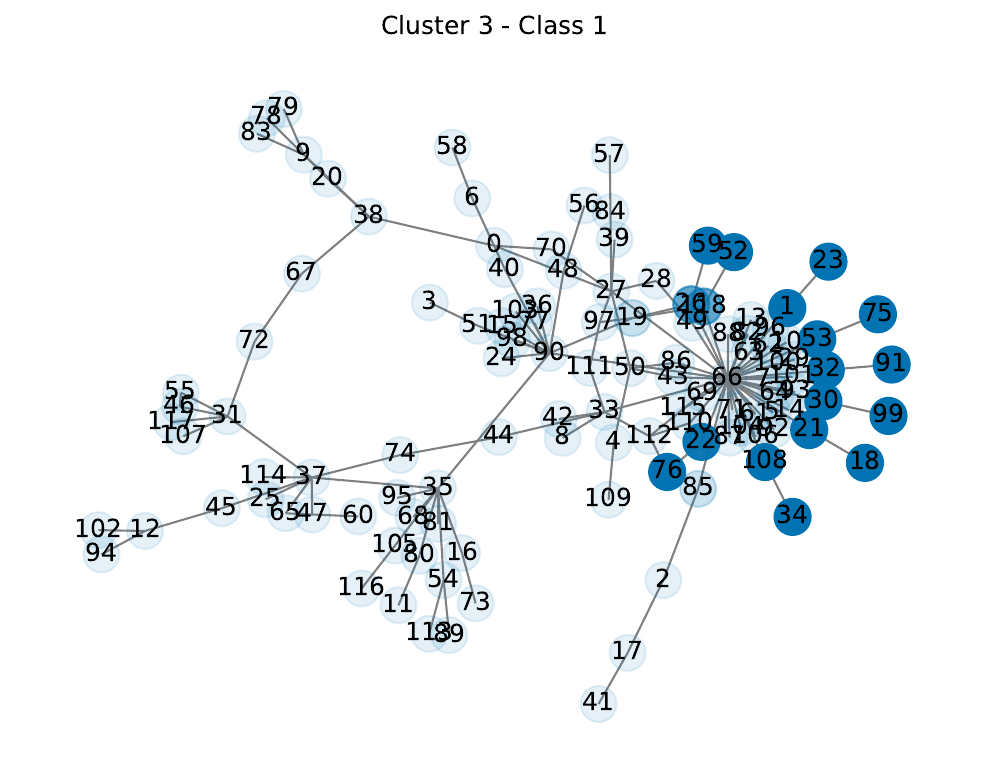}
            \label{fig:reddit_b}
        }

        \vspace{0.1em}

        \subfloat[Subgraph 5]{%
            \includegraphics[width=\textwidth]{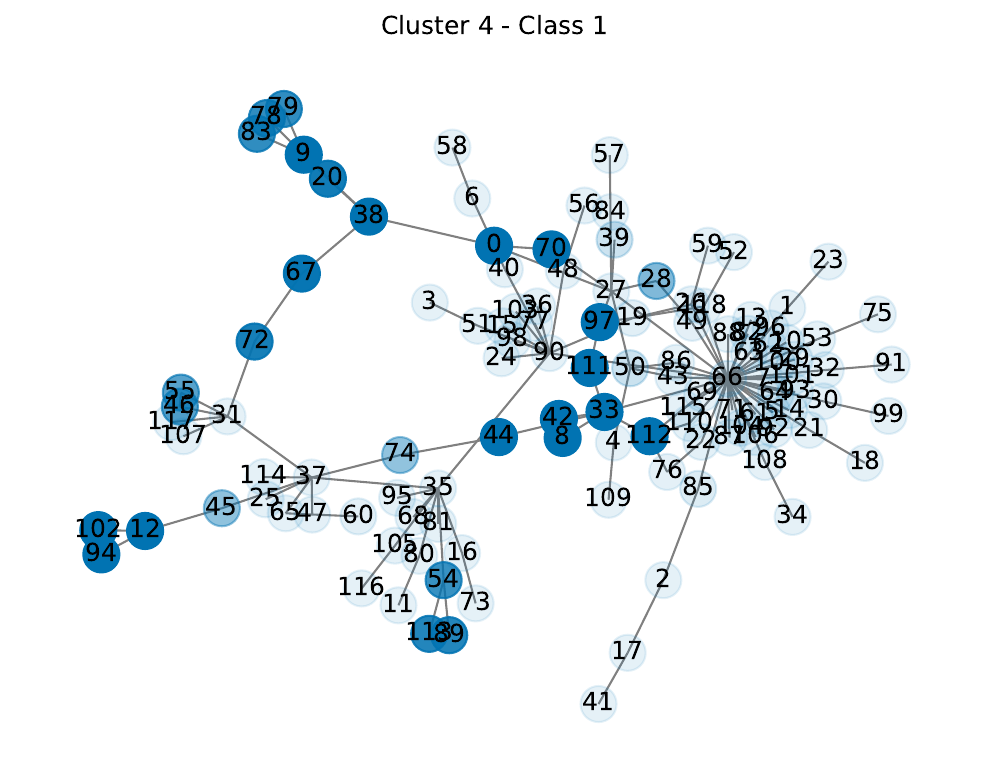}
            \label{fig:reddit_c}
        }
    \end{minipage}%
    \hspace{0.01\textwidth}
    \begin{minipage}[t]{0.3\textwidth}
        \vspace{0pt}
        \centering
        \subfloat[Subgraph 7]{%
            \includegraphics[width=\textwidth]{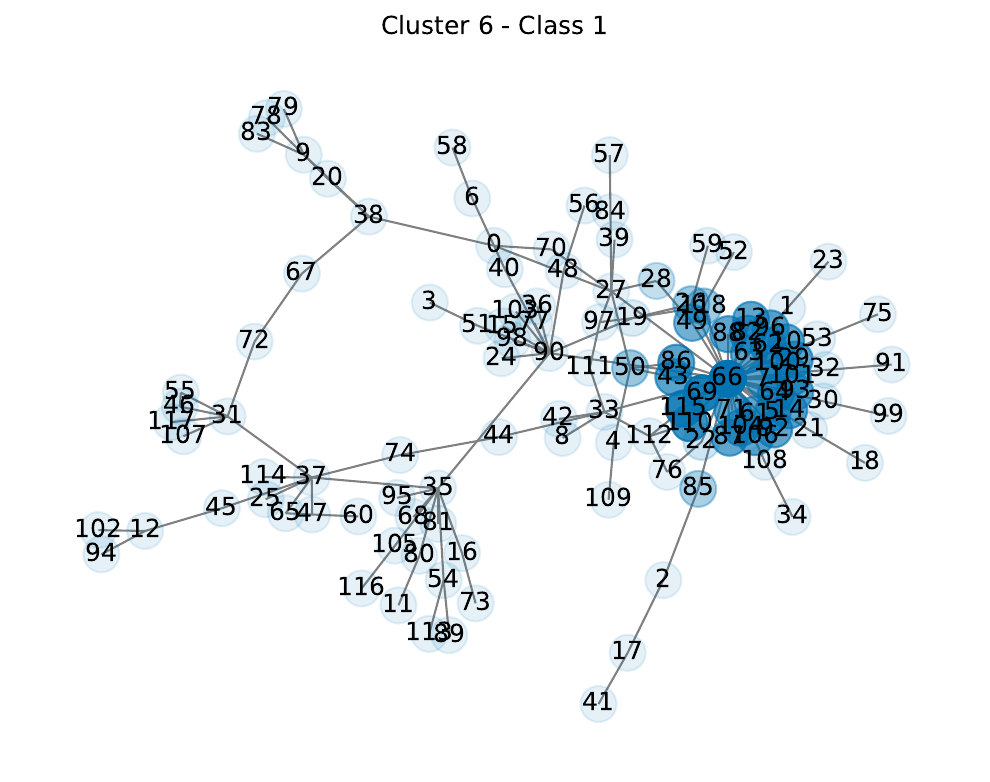}
            \label{fig:reddit_d}
        }

        \vspace{0.1em}

        \subfloat[Subgraph 10]{%
            \includegraphics[width=\textwidth]{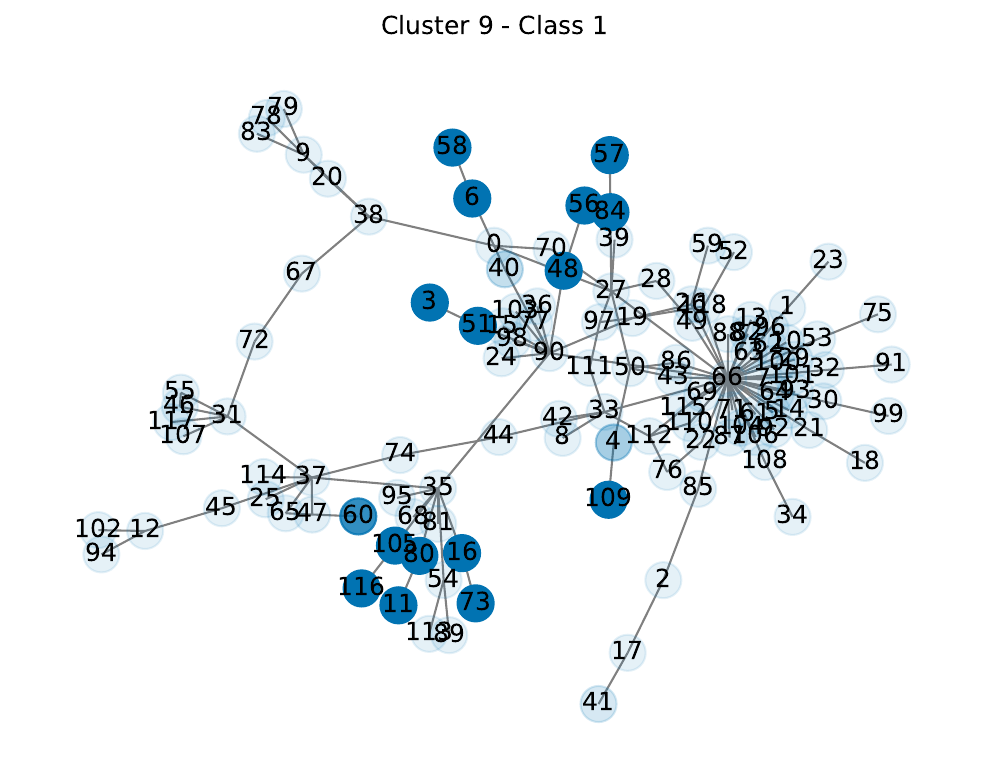}
            \label{fig:reddit_e}
        }
    \end{minipage}
    \caption{Instance-level explanation for the Reddit-Binary dataset, selecting 4 out of 10 subgraphs for visualisation. The range of subgraphs selected for visualisation shows that the Subgraph Concept Network successfully identifies the Q/A-structure in the dataset (Subgraphs 4 and 7), while also encoding the less highly connected base structure (Subgraphs 5 and 10). Examining the subgraph importance scores here, we can see that Subgraphs 5 and 10 receive a higher importance score than Subgraphs 4 and 7, indicating that the model focuses more on the less connected structures. This is in line with class 1 representing the discussion threads with lower engagement.}
    \label{fig:reddit_instance}
\end{figure}

Figure \ref{subgraph_concept_stars} visualises a concept extracted from the STARS dataset. The SCN correctly separate the base graph from the star structures. We can also visualize instance-level explanations for the SCN. For example, Figure \ref{fig:house_color_instance} explains a graph from the House-Colour dataset. We can see a clear separation of the desired motif, a green house structure. However, note that it appears as if the subgraph importance score for Subgraph 2 (Figure \ref{fig:house_color_c}), the predictive pattern, is the lowest at 0.03. This does not indicate that the model does not treat this as the most important subgraph. Rather, we do not enforce that importance must mean 1 in our architecture. This can be seen as a drawback, as a higher value indicating the most important subgraph is more intuitive. A range in the importance scores is also shown in Figure \ref{fig:reddit_instance} for the Reddit-Binary dataset. Nevertheless, we successfully identify the desired Q/A-structure (\ref{fig:reddit_b} and \ref{fig:reddit_d})). 

\begin{figure}[h]
\centerline{\includegraphics[width=0.9\textwidth]{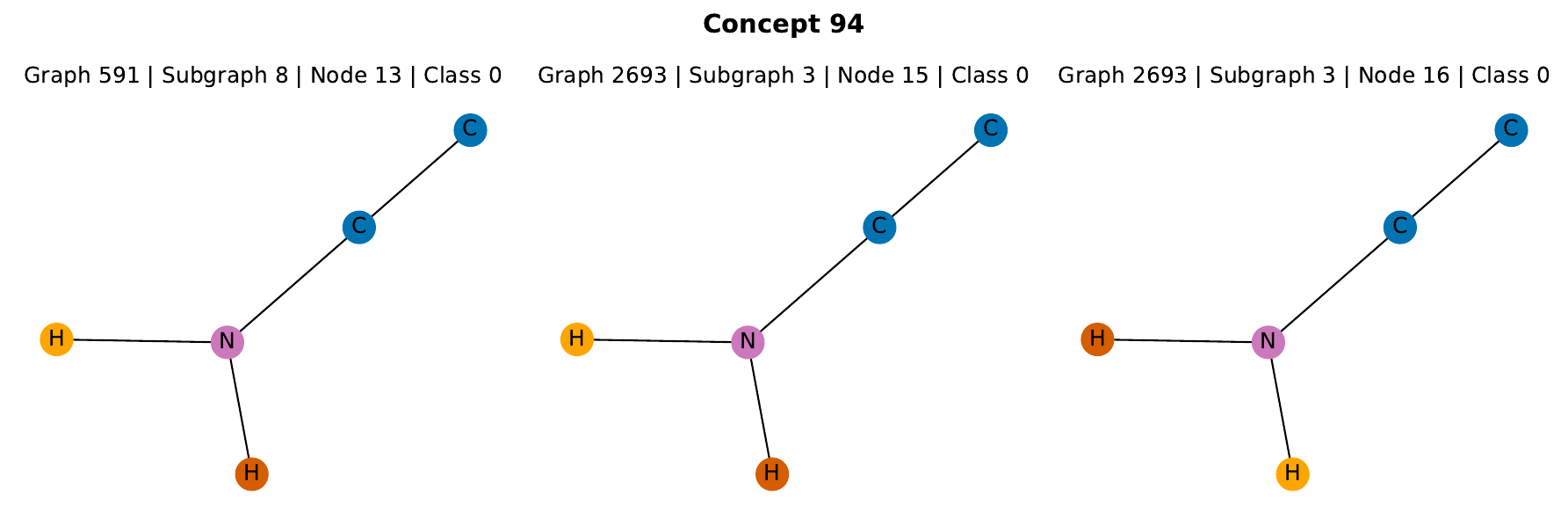}}
\caption{Node concepts discovered at the subgraph level for the Mutagenicity dataset, showing that the atom $H$ in a $NH_2$ structure is successfully identified. Notice that two out of three instances are assigned to the same clustering slot.}
\label{node_concept_mutag}
\end{figure}

\begin{figure}[h]
\centerline{\includegraphics[width=0.9\textwidth]{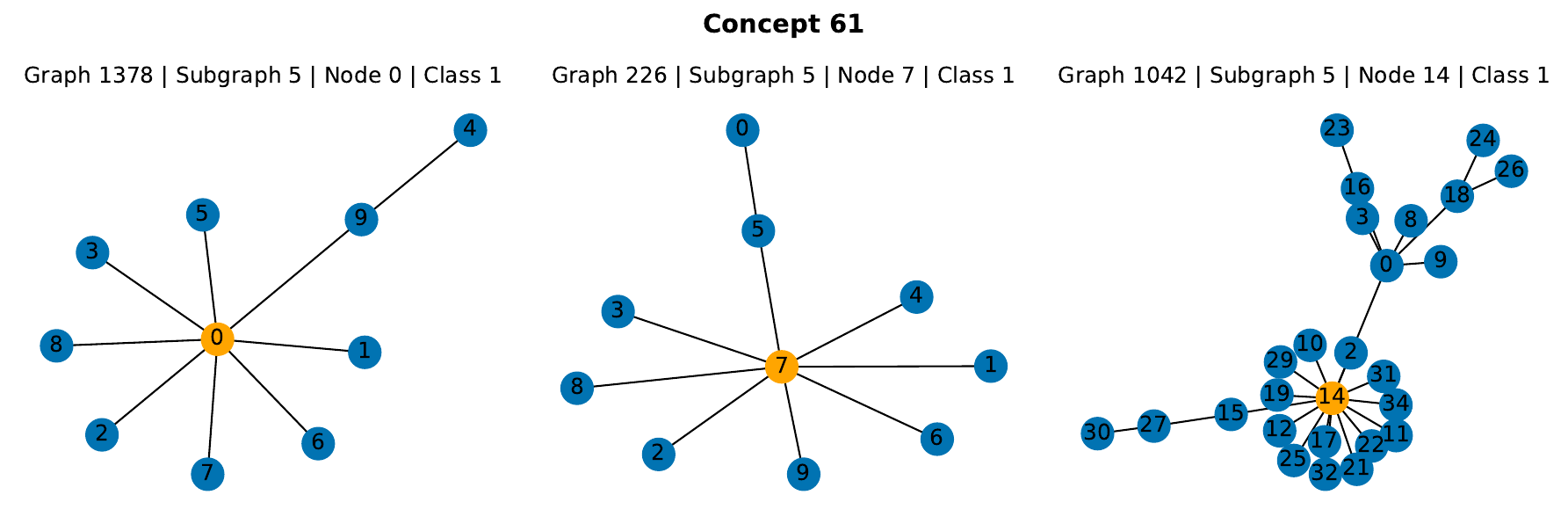}}
\caption{Node concepts discovered at the subgraph level for the Reddit-Binary dataset, showing that the model can successfully identify the central node of a star structure, representing a Q/A discussion thread.}
\label{node_concept_reddit}
\end{figure}

\paragraph{Node-level concept visualisations} We can also visualize node-level concepts for the SCN. For example, Figures \ref{node_concept_mutag} and \ref{node_concept_reddit} show the node-level concepts discovered for the Mutagenicity and Reddit-Binary datasets at the subgraph level, respectively. In Figure \ref{node_concept_mutag}, we can see that the network successfully identifies the atom $H$ in a $NH_2$ structure, while in Figure \ref{node_concept_reddit}, we can see that the central node of a star structure is identified, recovering the desired motif of a Q/A discussion thread.

\section{Analysis of cluster usage}
\label{analysis_cluster_usage}

\paragraph{Subgraph assignment strength} The subgraph assignment strength quantifies how strongly nodes in a graph are assigned to a cluster on average. Table \ref{fig:subgraph_assignment_strength} summarizes the subgraph assignment strengths computed for the SCN and CGN using DiffPool \citep{diffpool}. Overall, we find that the CGN with DiffPool exhibits a higher subgraph assignment strength, which means that nodes are assigned to a cluster more strongly. However, the SCN performs competitively on the Grid and STARS datasets. Both the SCN and CGN using DiffPool have a higher subgraph assignment strength than the lower bound of $\frac{1}{K}$, where $K$ is the number of clusters, indicating that on average nodes are assigned to a distinct cluster. Specifically, we find fairly strong assignment on the Grid and Grid-House dataset, given the respective threshold of $\frac{1}{K}$. We see fairly low subgraph assignment strength for the SCN on Mutagenicity and Reddit-Binary, which can be explained by the number of subgraph clusters potentially being too high at 10. While the subgraph assignment strength seems to indicate that SCN does not perform as well as the CGN with DiffPool, the score does not paint the full picture. Recall, that a flaw of the subgraph assignment strength metric is that it does not quantify how distributed the assignments of the nodes are across the remaining clusters. This is quantified by the conditional subgraph assignment strength and cluster utilization.

\begin{table*}[h]
\caption{Average subgraph assignment strength for the Subgraph Concept Network (SCN) and a Concept Graph Network (CGN) using DiffPool \citep{diffpool}, indicating how strongly nodes are assigned to their respective cluster. Overall, it appears as if the CGN with DiffPool exhibits higher cluster assignment strength.}
\centering
\resizebox{1\textwidth}{!}{
\begin{tabular}{lc|ll|ll}
\hline
 & \multicolumn{5}{c}{\textbf{\begin{tabular}[c]{@{}c@{}}Subgraph Assignment Strength\end{tabular}}} \\
 & \multirow{2}{*}{\textbf{$\frac{1}{K}$}} & \multicolumn{2}{c}{\textbf{SCN}} & \multicolumn{2}{c}{\textbf{CGN w/ DiffPool}} \\
 & & \multicolumn{1}{c}{\textbf{Train}} & \multicolumn{1}{c}{\textbf{Test}} & \multicolumn{1}{|c}{\textbf{Train}} & \multicolumn{1}{c}{\textbf{Test}}     \\ 
 \hline
\textbf{Grid} & 0.50 & 0.67 (0.65, 0.69) & 0.66 (0.64, 0.68) & \textbf{0.69 (0.69, 0.69)} & \textbf{0.68 (0.68, 0.68)} \\
\textbf{Grid-House} & 0.25 & 0.78 (0.76, 0.80) & 0.78 (0.76, 0.81) & \textbf{0.82 (0.82, 0.82)} & \textbf{0.82 (0.82, 0.82)} \\
\textbf{STARS} & 0.50 & \textbf{0.69 (0.68, 0.69)} & \textbf{0.69 (0.68, 0.69)} & 0.69 (0.69, 0.70) & 0.69 (0.69, 0.70) \\
\textbf{House-Colour} & 0.25 & 0.63 (0.62, 0.64) & 0.63 (0.61, 0.64) & \textbf{0.67 (0.67, 0.67)} & \textbf{0.68 (0.65, 0.68)} \\
\hline
\textbf{Mutagenicity} & 0.10 & 0.25 (0.25 0.25) & 0.25 (0.25, 0.25) & \textbf{0.29 (0.29, 0.29)} & \textbf{0.29 (0.29, 0.29)} \\
\textbf{Reddit-Binary} & 0.10 & 0.33 (0.31, 0.35) & 0.32 (0.29, 0.36) & \textbf{0.44 (0.44, 0.45)} & \textbf{0.44 (0.41, 0.46)} \\
\hline
\end{tabular}%
}

\label{fig:subgraph_assignment_strength}
\end{table*}

\paragraph{Conditional subgraph cluster assignment strength} To analyse the cluster utilization further, Table \ref{fig:individual_conditional_strengths} provides the conditional subgraph cluster assignments per cluster for the SCN on all datasets. For the synthetic datasets, we can observe that across all clusters, the conditional subgraph assignment strength is fairly stable. In comparison, we observe more fluctuations across the clusters in the SCN trained on the Mutagenicity and Reddit-Binary datasets. For example, referring to the scores for the Reddit-Binary dataset, the lowest subgraph assignment strength is 0.39 for cluster 1, while the highest is 0.57 for cluster 10. This indicates that potentially not all clusters are needed. This stands in tension with the utilisation loss term, which encourages not all nodes to be assigned to a single cluster. This lets us hypothesize that examining these values can be useful in deciding the parametrization needs of the model for a given dataset. Overall, we can say that all clusters in the SCN are used across these datasets. In contrast, Table \ref{fig:individual_conditional_strengths2} shows the conditional subgraph assignment strengths per cluster for the CGN using DiffPool \citep{diffpool}. We can observe that the conditional cluster assignment strength fluctuates much more. For example, in the model trained on the STARS dataset, all nodes are assigned to the same cluster, indicating a lack of identifying a clustering in the nodes. Combined with the competitive performance of the model on the STARS dataset, this indicates a lack of interpretability via clustering due to a lack of additional loss terms helping the disentanglement of the latent space.

\begin{table*}[ht]
\caption{Conditional subgraph assignment scores of the Subgraph Concept Network (SCN), measuring the average cluster assignment strength for each cluster, where a node is deemed part of a cluster if the activation exceed $\frac{1}{K}$, where $K$ is the number of clusters.}
\centering
\resizebox{1.0\textwidth}{!}{
\begin{tabular}{ccccccc}
\hline
\multicolumn{7}{c}{\textbf{Conditional Subgraph Assignment Strength for the SCN on the Test Slice}} \\
\textbf{Cluster} & \textbf{Grid} & \textbf{Grid-House} & \textbf{STARS} & \textbf{House-Colour} & \textbf{Mutagenicity} & \textbf{Reddit-Binary} \\
\hline
\textbf{1}  & 0.96 (0.94, 0.99) & 0.92 (0.86, 0.98) & 0.99 (0.99, 0.99) & 0.89 (0.86, 0.92) & 0.71 (0.62, 0.80) & 0.39 (0.22, 0.55) \\
\textbf{2}  & 0.97 (0.96, 0.98) & 0.92 (0.86, 0.97) & 0.99 (0.98, 0.99) & 0.89 (0.87, 0.91) & 0.70 (0.62, 0.78) & 0.40 (0.24, 0.56) \\
\textbf{3}  & - & 0.93 (0.89, 0.98) & - & 0.90 (0.88, 0.91) & 0.65 (0.56, 0.74) & 0.45 (0.30, 0.60) \\
\textbf{4}  & - & 0.88 (0.77, 1.00) & - & 0.90 (0.85, 0.95) & 0.72 (0.65, 0.78) & 0.52 (0.35, 0.69) \\
\textbf{5}  & - & - & - & - & 0.70 (0.63, 0.76) & 0.46 (0.28, 0.63) \\
\textbf{6}  & - & - & - & - & 0.63 (0.49, 0.78) & 0.43 (0.28, 0.58) \\
\textbf{7}  & - & - & - & - & 0.67 (0.58, 0.76) & 0.43 (0.26, 0.60) \\
\textbf{8}  & - & - & - & - & 0.67 (0.56, 0.78) & 0.44 (0.27, 0.61) \\
\textbf{9}  & - & - & - & - & 0.70 (0.65, 0.76) & 0.54 (0.38, 0.70) \\
\textbf{10} & - & - & - & - & 0.67 (0.55, 0.80) & 0.57 (0.39, 0.74)\\
\hline
\end{tabular}
}

\label{fig:individual_conditional_strengths}
\end{table*}

\begin{table*}[ht]
\caption{Conditional subgraph assignment scores of the Concept Graph Network (CGN) using DiffPool \citep{diffpool}, measuring the average cluster assignment strength for each cluster, where a node is deemed part of a cluster if the activation exceed $\frac{1}{K}$, where $K$ is the number of clusters.}
\centering
\resizebox{1.0\textwidth}{!}{
\begin{tabular}{ccccccc}
\hline
\multicolumn{7}{c}{\textbf{Conditional Subgraph Assignment Strength for the CGN w/ DiffPool on the Test Slice}} \\
\textbf{Cluster} & \textbf{Grid} & \textbf{Grid-House} & \textbf{STARS} & \textbf{House-Colour} & \textbf{Mutagenicity} & \textbf{Reddit-Binary} \\
\hline
\textbf{1}  & 0.80 (0.25, 1.00) & 0.40 (0.00, 1.00) & 1.00 (1.00, 1.00) & 0.20 (0.00, 0.75) & 0.40 (0.15, 0.65) & 0.29 (0.15, 0.43) \\
\textbf{2}  & 0.20 (0.00, 0.76) & 0.00 (0.00, 0.00) & 0.00 (0.00, 0.00) & 0.60 (0.00, 1.00) & 0.37 (0.12, 0.62) & 0.40 (0.23, 0.56) \\
\textbf{3}  & - & 0.20 (0.00, 0.76) & - & 0.20 (0.00, 0.75) & 0.41 (0.19, 0.63) & 0.34 (0.19, 0.48) \\
\textbf{4}  & - & 0.40 (0.00, 1.00) & - & 0.00 (0.00, 0.00) & 0.51 (0.26, 0.76) & 0.39 (0.23, 0.55) \\
\textbf{5}  & - & - & - & - & 0.37 (0.12, 0.62) & 0.39 (0.23, 0.56) \\
\textbf{6}  & - & - & - & - & 0.36 (0.14, 0.59) & 0.32 (0.18, 0.47) \\
\textbf{7}  & - & - & - & - & 0.46 (0.20, 0.72) & 0.37 (0.21, 0.53) \\
\textbf{8}  & - & - & - & - & 0.45 (0.30, 0.69) & 0.33 (0.17, 0.48) \\
\textbf{9}  & - & - & - & - & 0.50 (0.25, 0.76) & 0.40 (0.24, 0.56) \\
\textbf{10} & - & - & - & - & 0.55 (0.27, 0.82) & 0.47 (0.30, 0.65) \\
\hline
\end{tabular}
}

\label{fig:individual_conditional_strengths2}
\end{table*} 

\paragraph{Subgraph consistency score} Finally, we can inspect the subgraph consistency score to assess the performance of our additional loss terms. Table \ref{fig:consistency_score} collates the subgraph consistency scores computed for the SCN and CGN with DiffPool \citep{diffpool}. The Concept Graph Network (CGN) using DiffPool achieves a subgraph consistency score of $1.0$ across all datasets. This means that all nodes with the same concept embedding are assigned to the same cluster across graphs. We highlight this score in grey, becasue when examining the the conditional subgraph assignment, we can observe that not all clusters are used. More specifically, in some cases only 1 cluster is used, such as for the STARS dataset. This makes it trivially easy to obtain a subgraph consistency score of $1.00$. Moreover, across the Grid and House-Colour datasets, we can identify clear clusters to which the nodes would be assigned under our metric, leading to the perfect score of $1.00$. Given this, we identify the SCN to perform better here. Given the cluster usage, we can explain the subgraph consistency score not being $1.00$ by some nodes being assigned to different clusters either because they are not important for the task or appear multiple times in the task. Furthermore, this could be caused by the connectivity loss, which encourages connected nodes to be assigned to the same graph. This means that nodes at the boundary of two subgraphs are harder to assign to one cluster or another. Our analysis of the visualisations shows that the SCN is able to recover graph structures important for prediction via the concept-based soft clustering.

\begin{table*}[ht]
\caption{Average subgraph consistency of the Subgraph Concept Network (SCN) and a CGN using DiffPool \citep{diffpool}, quantifying whether node-level concepts are consistently assigned to the same cluster across graphs. We bold the metric of the best performing model and grey out the invalid scores of $1.0$, where all nodes are assigned to the same cluster, trivially leading to this score.}
\centering
\resizebox{0.8\textwidth}{!}{
\begin{tabular}{lll|ll}
\hline
 & \multicolumn{4}{c}{\textbf{\begin{tabular}[c]{@{}c@{}}Subgraph Consistency\end{tabular}}} \\
 & \multicolumn{2}{c}{\textbf{SCN}} & \multicolumn{2}{c}{\textbf{CGN w/ DiffPool}} \\
 & \multicolumn{1}{c}{\textbf{Train}} & \multicolumn{1}{c}{\textbf{Test}} & \multicolumn{1}{|c}{\textbf{Train}} & \multicolumn{1}{c}{\textbf{Test}}     \\ 
 \hline
\textbf{Grid} & \textbf{0.85 (0.70, 1.00)} & \textbf{0.79 (0.60, 0.98)} & \textcolor{gray}{1.00 (1.00, 1.00)} & \textcolor{gray}{1.00 (1.00, 1.00)} \\
\textbf{Grid-House} & \textbf{0.84 (0.76, 0.91)} & \textbf{0.81 (0.72, 0.90)} & \textcolor{gray}{1.00 (1.00, 1.00)}  & \textcolor{gray}{1.00 (1.00, 1.00)} \\
\textbf{STARS} & \textbf{0.72 (0.50, 0.95)} & \textbf{0.76 (0.59, 0.94)} & \textcolor{gray}{1.00 (1.00, 1.00)} & \textcolor{gray}{1.00 (1.00, 1.00)} \\
\textbf{House-Colour} & \textbf{0.92 (0.85, 0.90)} & \textbf{0.89 (0.79, 0.98)} & \textcolor{gray}{1.00 (1.00, 1.00)} & \textcolor{gray}{1.00 (1.00, 1.00)} \\
\hline
\textbf{Mutagenicity} & \textbf{0.73 (0.68, 0.77)} & \textbf{0.73 (0.69, 0.77)} & \textcolor{gray}{1.00 (1.00, 1.00)} & \textcolor{gray}{1.00 (1.00, 1.00)} \\
\textbf{Reddit-Binary} & \textbf{0.68 (0.63, 0.73)} & \textbf{0.67 (0.62, 0.73)} & \textcolor{gray}{1.00 (1.00, 1.00)} & \textcolor{gray}{1.00 (1.00, 1.00)} \\
\hline
\end{tabular}%
}

\label{fig:consistency_score}
\end{table*}

\section{Qualitative comparison to graph interactive pattern learning}
\label{qualitative_gip}

\begin{figure}[h]
\centerline{\includegraphics[width=0.8\textwidth]{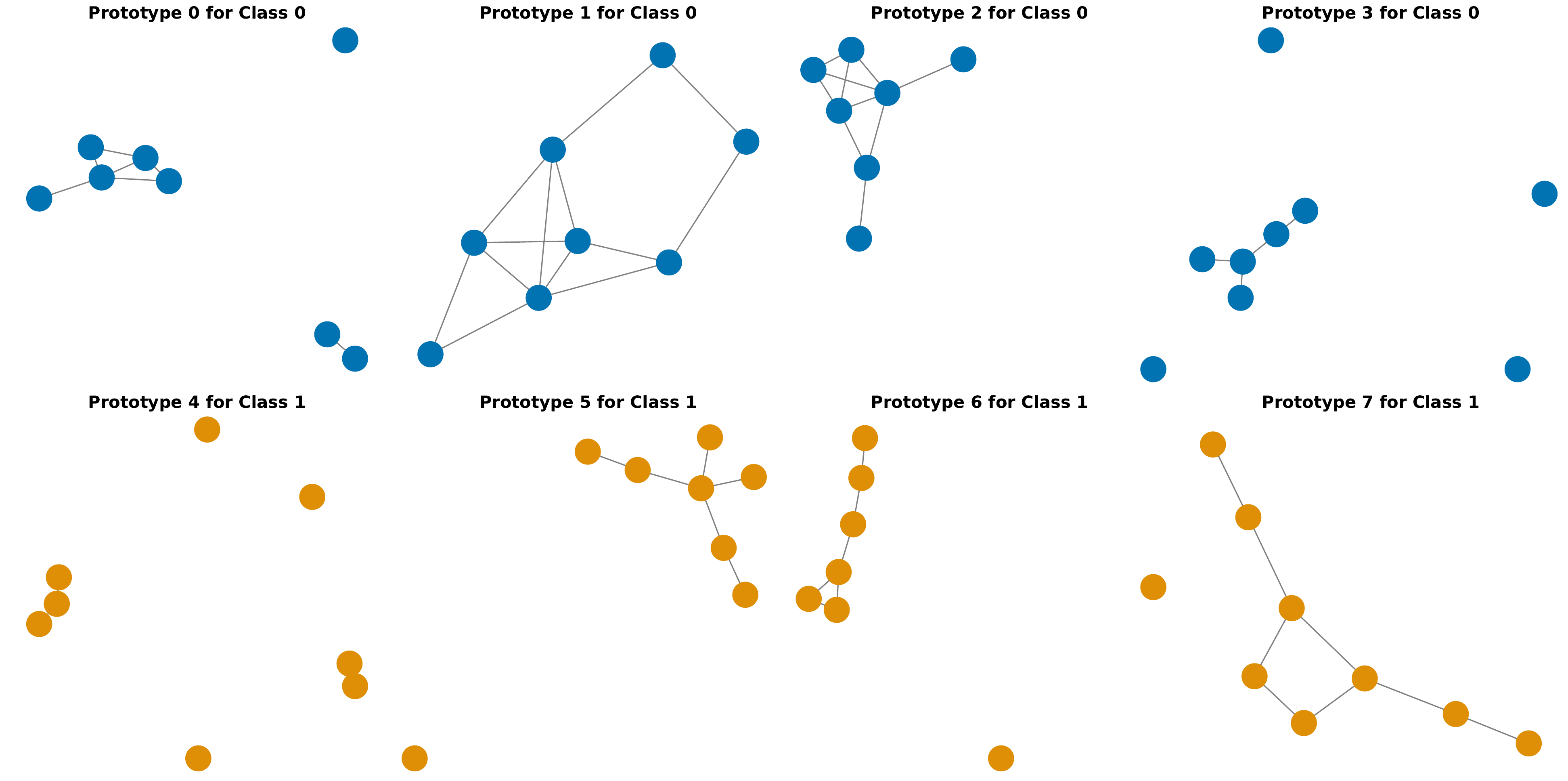}}
\caption{Prototypes discovered with the GIP method \citep{wang2024unveiling} for the Grid dataset. Four patterns per class are discovered. We cannot identify a grid structure among the classes, however, this might be due to the parameterization of our method being set similar to that of the Subgraph Concept Network for comparison purposes.}
\label{gip_grid}
\end{figure}

\begin{figure}[h]
\centerline{\includegraphics[width=0.8\textwidth]{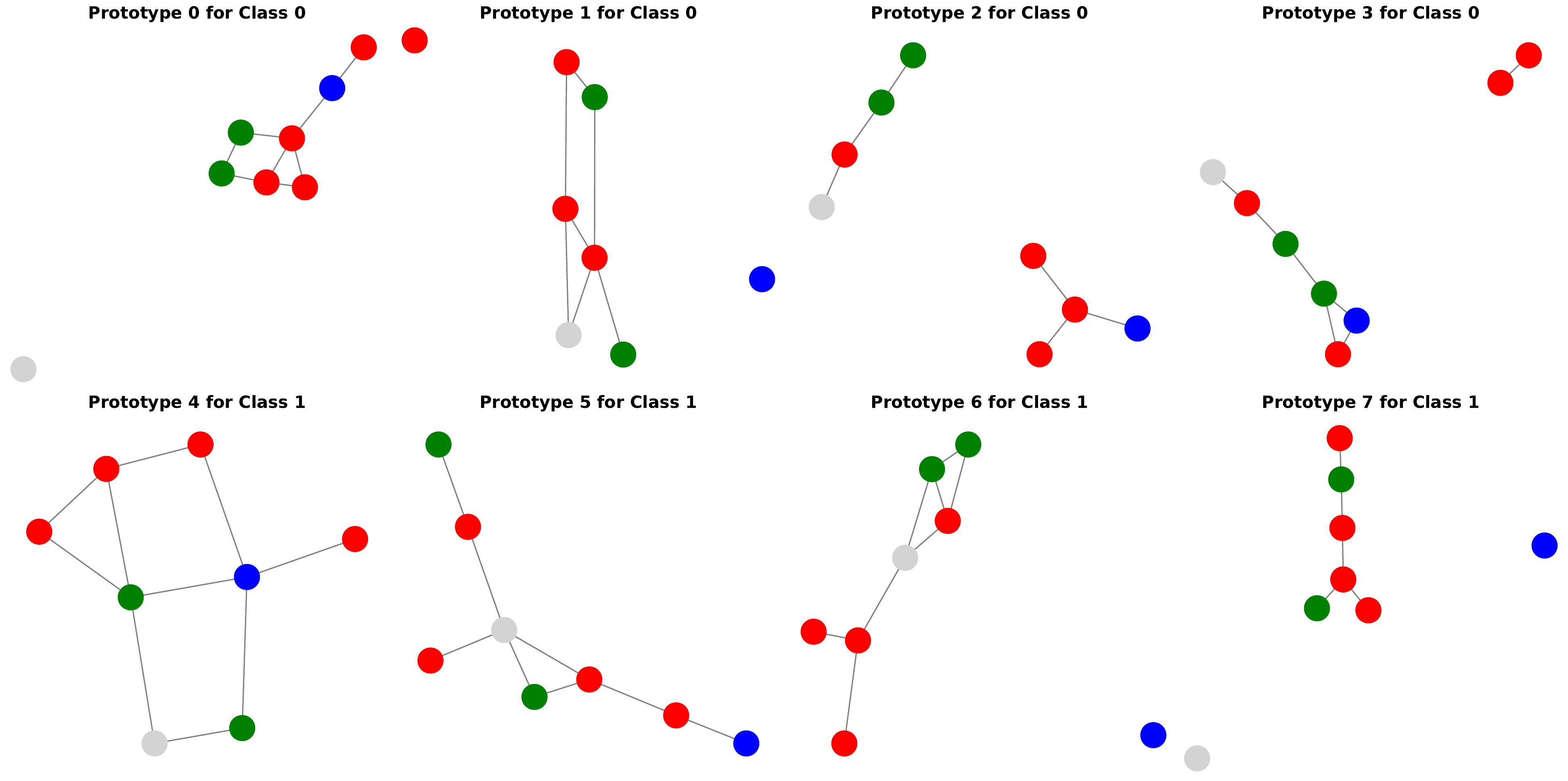}}
\caption{Prototypes discovered with the GIP \citep{wang2024unveiling} method for the House-Colour dataset, with four prototypes per class. We can identify prototypes with the house structure across both classes. Note, that the node colouring is an addition we add since the feature labels are important for this task.}
\label{grip_house_colour}
\end{figure}

\begin{figure}[h]
\centerline{\includegraphics[width=1\textwidth]{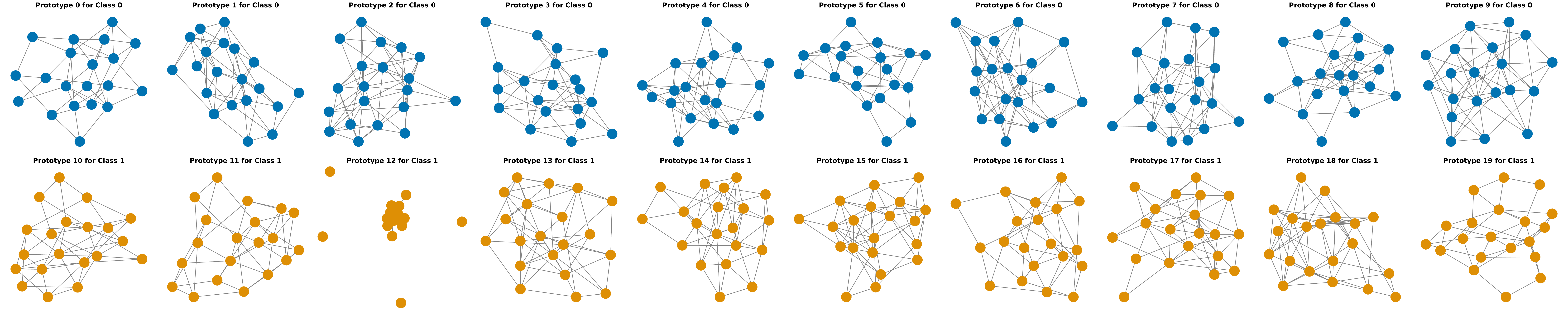}}
\caption{Prototypes discovered with the GIP \citep{wang2024unveiling} method for the Reddit-binary dataset, with ten prototypes per class. Overall, the prototypes for each class look similar, which may indicate that we set the number of prototypes too high. However, we can identify densely connected structures, which is indicative for the classes of the dataset.}
\label{gip_reddit}
\end{figure}

We can broadly compare the explanations extracted from the GIP method \citep{wang2024unveiling} to the explanations extracted from the SCN. The visual explanations for the GIP method are the learned prototypes. Figures \ref{gip_grid}, \ref{grip_house_colour} and \ref{gip_reddit} show the prototypes for the Grid, House-Colour and Reddit-Binary dataset, respectively. Overall, we can say that we do not achieve to extract the desired motifs or produce visualisations as detailed as the ones for the SCN. However, this might stem from a wrong parametrization of the GIP model. We set the hyperparameters of the GIP method to match those of the SCN for equal expressivity, however, this might be the wrong parametrization. Specifically, we set the number of prototypes to the same number of subgraphs as in the SCN, however, the SCN does not necessarily strictly split subgraphs by connectivity, allowing to encode more motifs per subgraph than the GIP model. One drawback of the GIP method we observe is that the prototypes are simply the substructure but do not give a clear insight about the nodes, which is needed for datasets like the House-Colour or Mutagenicity. We extend the visualisation technique to majority vote how often the prototype-matched subgraphs have specific node labels and colour the graph based on this. However, this is only a rough approximation, highlighting the difference between prototype-based and concept-based explanations.

\end{document}